\def\year{2022}\relax
\documentclass[letterpaper]{article} 
\usepackage{aaai22}  
\usepackage{times}  
\usepackage{helvet}  
\usepackage{courier}  
\usepackage[hyphens]{url}  
\usepackage{graphicx} 
\usepackage{natbib}  
\usepackage{caption} 
\DeclareCaptionStyle{ruled}{labelfont=normalfont,labelsep=colon,strut=off} 
\usepackage{algorithm}
\usepackage{algorithmic}

\usepackage{cuted}
\usepackage{tabularx, booktabs} 
\usepackage{tikz}
\usepackage{pgfplots}
\usepackage{times}
\usepackage{epsfig}
\usepackage{graphicx}
\usepackage{amsmath}
\usepackage{amssymb}
\usepackage{xcolor}
\usepackage{amsfonts}
\usepackage{paralist}
\usepackage[tight,footnotesize,sf,SF]{subfigure}
\definecolor{ForestGreen}{rgb}{0.13, 0.55, 0.13}
\usepackage{pifont}%
\newcommand{\cmark}{{\ding{51}}}%
\newcommand{\xmark}{{ \ding{55}}}%

\usepackage{enumitem}
\setlist{topsep=0pt, leftmargin=*,noitemsep,topsep=0pt,parsep=0pt,partopsep=0pt}

\renewcommand{\subsection}[1]{\textbf{#1.}}
\renewcommand{\subsubsection}[1]{\textit{#1.}}
\usepackage{newfloat}
\usepackage{listings} 
\floatstyle{ruled}
\newfloat{listing}{tb}{lst}{}
\floatname{listing}{Listing}

\title{Joint Deep Multi-Graph Matching and 3D Geometry Learning \\from Inhomogeneous 2D Image Collections}
\author{

    Zhenzhang Ye\textsuperscript{\rm 1},
    Tarun Yenamandra\textsuperscript{\rm 1},
    Florian Bernard\textsuperscript{\rm 1, 2},
    Daniel Cremers\textsuperscript{\rm 1}
}
\affiliations{
    \textsuperscript{\rm 1}Technical University of Munich \\
    \textsuperscript{\rm 2}University of Bonn \\
    zhenzhang.ye@tum.de, tarun.yenamandra@tum.de, f.bernard@gmail.com, cremers@tum.de
}

\definecolor{green(ryb)}{rgb}{0.4, 0.69, 0.2}

\newtheorem{theorem}{Theorem}
\newtheorem{corollary}[theorem]{Corollary}
\newtheorem{lemma}[theorem]{Lemma}
\newtheorem{proposition}[theorem]{Proposition}
\newtheorem{definition}[theorem]{Definition}

\newcommand{\bthm}{\begin{theorem}}
\newcommand{\ethm}{\end{theorem}}
\newcommand{\bcor}{\begin{corollary}}
\newcommand{\ecor}{\end{corollary}}
\newcommand{\blem}{\begin{lemma}}
\newcommand{\elem}{\end{lemma}}
\newcommand{\bprop}{\begin{proposition}}
\newcommand{\eprop}{\end{proposition}}
\newcommand{\bdefn}{\begin{definition}}
\newcommand{\edefn}{\end{definition}}
\newcommand{\bpf}{\begin{proof}}
\newcommand{\epf}{\end{proof}}

\DeclareMathOperator{\argmin}{arg\,min}
\DeclareMathOperator{\argmax}{arg\,max}

\DeclareMathOperator{\diag}{diag}

\newcommand{\bR}{\mathbb{R}}

\newcommand{\bP}{\mathbb{P}}

\newcommand{\cL}{\mathcal{L}}
\newcommand{\cX}{\mathcal{X}}

\newcommand{\cE}{\mathcal{E}}

\newcommand{\cV}{\mathcal{V}}
\newcommand{\cU}{\mathcal{U}}	
\newcommand{\cG}{\mathcal{G}}

\begin{document}
\newcommand{\teaserheight}[0]{4cm}
\maketitle%

\begin{abstract}
Graph matching aims to establish correspondences between vertices of graphs such that both the node and edge attributes agree. Various learning-based methods were recently proposed for finding correspondences between image key points based on deep graph matching formulations. While these approaches mainly focus on learning node and edge attributes, they completely ignore the 3D geometry of the underlying 3D objects depicted in the 2D images. We fill this gap by proposing a trainable framework that takes advantage of graph neural networks for learning a deformable 3D geometry model from inhomogeneous image collections, i.e.,~a set of images that depict different instances of objects from the same category. Experimentally, we demonstrate that our method outperforms recent learning-based approaches for graph matching considering both accuracy and cycle-consistency error, while we in addition obtain the underlying 3D geometry of the objects depicted in the 2D images. 
\end{abstract}

\section*{Introduction}
Graph matching is a widely studied problem in computer vision, graphics and machine learning due to its universal nature and the broad range of applications. Intuitively, the objective of graph matching is to establish correspondences between the nodes of two given weighted graphs, so that the weights of corresponding edges agree as well as possible.
Diverse visual tasks fit into the graph matching framework. In this work we focus in particular on the task of matching 2D key points defined in images, which has a high relevance for 3D reconstruction, tracking, deformation model learning, and many more. In this case, a graph is constructed for each image by using the key points as  graph nodes, and by connecting neighbouring key points with edges, according to some suitable neighbourhood criterion. The edges contain information about geometric relations, such as the Euclidean distance between nodes in the simplest case.

Image key point matching was traditionally addressed based on finding nearest neighbours between feature descriptors such  as SIFT~\cite{lowe2004sift}, SURF~\cite{bay2008surf}.
A downside to this approach is that the geometric relation between the key points are completely ignored, which is in particular problematic if there are repetitive structures that lead to similar feature descriptors. 
Instead, we can use a graph matching formulation to establish correspondences between key points while taking into account geometric relations between points. 
Yet, the sequential nature of first computing features and then bringing them into correspondence may lead to sub-optimal results, since both tasks are solved independently from each other -- despite their mutual dependence. More recently, several deep learning-based graph matching methods have been proposed that learn task-specific optimal features while simultaneously solving graph matching in an end-to-end manner~\cite{zanfir2018gmn,wang2019PCA,wang2020combinatorial,rolinek2020deep}. While such deep graph matching approaches lead to  state-of-the-art results in terms of the matching accuracy, they have  profound disadvantages, particularly in the context of 2D key point matching in image collections. On the one hand, most existing approaches only consider the matching of pairs of images, rather than the entire collection. This has the negative side-effect that so-obtained matchings are generally not cycle-consistent.
To circumvent this, there are approaches that use a post-processing procedure
~\cite{wang2019neural}
to establish cycle consistency based on permutation synchronisation~\cite{pachauri2013permsync,bernard2018nmfsync}.
Yet, they do not directly obtain cycle-consistent matchings but rather achieve it based on post-processing. On the other hand, and perhaps more importantly, approaches that use graph matching for 2D image key point matching have the strong disadvantage that the underlying 3D structure of the objects whose 2D projections are depicted in the images is not adequately considered. In particular, the spatial relations in the 2D image plane are highly dependent on the 3D geometric structure of the object, as well as on the camera parameters. Hence, learning graph features directly based on the image appearance and/or 2D image coordinates is sub-optimal, at best, since the neural network implicitly needs to learn the difficult task of reasoning about the underlying 3D structure.
\begin{figure*}%
	\centerline{%
		\footnotesize%
		\begin{tabular}{cc|c}%
			\includegraphics[height=\teaserheight]{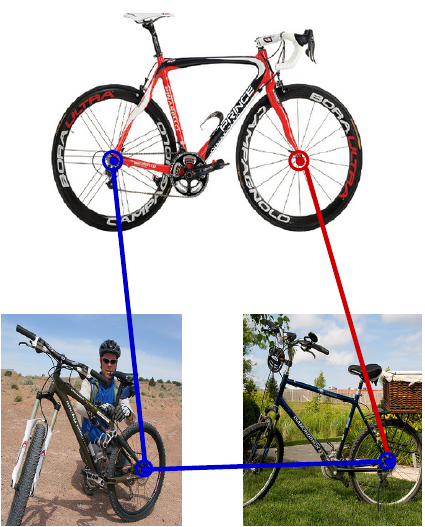} ~~&~~
			\includegraphics[height=\teaserheight]{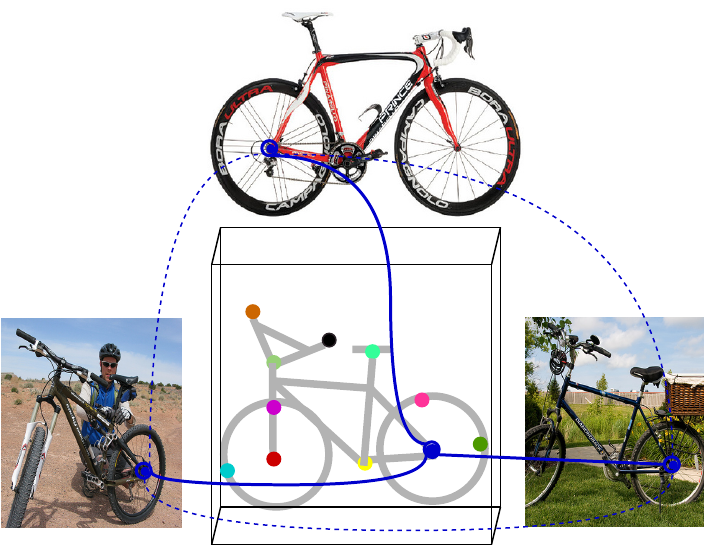}~~ & ~~
			\includegraphics[height=\teaserheight]{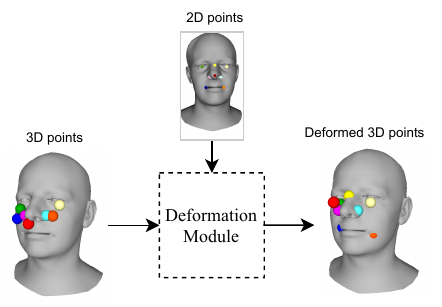}\\
			Previous approach~\cite{wang2020combinatorial} & \multicolumn{1}{c}{Our proposed approach} & \multicolumn{1}{c}{Sparse non-rigid geometry 3D reconstruction}
		\end{tabular}%
		
	}%
	\captionof{figure}{We consider a deep graph matching approach for bringing 2D image key points into correspondence. Left: Existing deep graph matching methods completely ignore the underlying 3D geometry of the 3D objects depicted in the 2D images. In addition, they lead to cycle errors, as shown by the red line. Middle: Our method obtains the underlying 3D geometry from a collection of inhomogeneous 2D images (indicated by the coloured points and the bike sketch in the centre), while at the same time guaranteeing cycle consistency. Right: To model nonlinear 3D object deformations, we infer coarse 3D geometry and in addition use a 3D deformation module to refine the underlying 3D geometry based on the 2D image key point observations.}
	\label{fig:teaser}
\end{figure*}

In this work we address these issues by proposing a deep multi-graph matching approach that learns the 3D structure of objects. The main contributions are as follows:

\begin{compactitem}
    \item For the first time we propose a solution for jointly considering multi-graph matching and inferring 3D geometry from inhomogeneous 2D image collections, see Fig.~\ref{fig:teaser}.
    \item To effectively deal with the inhomogeneity of the image collection, in which different instances of objects of the same category are present (e.g.~different types of bikes as shown in Fig.~\ref{fig:teaser}),
    we introduce a novel  deformable 3D model that we directly learn from the image collection based on a graph neural network.
     \item Rather than performing pairwise image-to-image matching, we consider an image-to-deformable-3D-model matching formulation to guarantee cycle consistency.
    \item Our approach substantially outperforms the previous state of the art in learning-based graph matching approaches considering accuracy and cycle error.
\end{compactitem}
\section*{Related Work}
In the following we summarise the works that we consider most relevant to our approach. For a more detailed background on image key point matching we refer interested readers to the recent survey paper by \citet{ma2021image}.

\textbf{Feature-Based Matching.} Feature descriptors extracted from images at key point locations,~e.g.~based on SIFT~\cite{lowe2004sift}, SURF~\cite{bay2008surf}, or deep neural networks~\cite{krizhevsky2012alexnet}, are often used for image matching. In order to bring extracted features into correspondence, commonly a nearest neighbour strategy~\cite{bentley1975knn} or a  linear assignment problem (LAP) formulation are used~\cite{burkard2012}.
However, these methods suffer from the problem that
geometric relations between the key points in the images are not taken into account.

\textbf{Graph Matching and Geometric Consistency.}  Geometric relations can be taken into account by modelling feature matching as graph matching problem. Here, the image key points represent the graph nodes, and the edges in the graph encode geometric relations between key points (e.g.~spatial distances).
Mathematically, graph matching can be phrased in terms of the quadratic assignment problem~\cite{lawler1963qap,pardalos1994quadratic,loiola2007survey,burkard2012}.
There are many existing works for addressing the graph matching problem in visual computing, including~\citet{Cour:2006un,Zhou:2016ty,swoboda2017b,dym2017dspp,bernard2018dsstar,swoboda2017b}. 
A drawback of these approaches is that they mostly rely on handcrafted graph attributes and/or respective graph matching cost functions based on affinity scores.
In~\citet{zhang2013Oriented}, a learning-based approach that directly obtains affinity scores from data was introduced. 
The differentiation of the power iteration method has been considered in a deep graph matching approach~\cite{zanfir2018gmn}.
A more general blackbox differentiation approach was introduced by \citet{rolinek2020deep}.
Various other deep learning approaches have been proposed for graph matching~\cite{li2019graph,fey2020deep}, and some approaches also address image key point matching~\cite{wang2019PCA,zhang2019deep,wang2020combinatorial}. In this case, optimal graph features are directly learned from the image appearance and/or 2D image coordinates, while simultaneously solving graph matching in an end-to-end manner.
Although these methods consider geometric consistency, they are tailored towards matching a pair of graphs and thus lead to cycle-inconsistent matchings when pairwise matchings of more than two graphs are computed.

\textbf{Synchronisation and Multi-Matching.} Cycle-consistency is often obtained as a post-processing step after obtaining pairwise matchings. The procedure to establish cycle consistency in the set of pairwise matchings is commonly referred to as permutation synchronisation~\cite{pachauri2013permsync,zhou2015multi,Maset:YO8y6VRb,bernard2018nmfsync,birdal2019probabilistic,bernard2021neurips}.
There are also methods for directly obtaining cycle-consistent multi-matchings~\cite{Tron17,wang2018,bernard19}. Recently, permutation synchronisation has been considered in a deep graph matching framework, where a separate permutation synchronisation module is utilised to generalise a two-graph matching approach to the matching of multiple graphs~\cite{wang2019neural}.
However, when applying such multi-matching approaches to image key point matching they have the significant shortcoming that they ignore the underlying 3D geometry of the 2D points.  This makes it extremely difficult to establish correct matchings across images, which after all depict 2D projections of 3D objects in different poses, possibly even under varying perspective projections. This also applies to the recent method by \citet{wang2020Advances}, which simultaneously considers  graph matching and clustering. 

\textbf{3D Reconstruction.}
3D reconstruction obtains geometric information from 2D data. When relying on single-view input only, it is generally an ill-posed problem.
Reconstruction from a single image or video using a deformable 3D prior has for example been achieved by fitting a 3D morphable model of a specific object class such as humans bodies, faces, or cars, and then finding the parameters of the model that best explain the image~\cite{tewari17MoFA,bogo2016keepitsmpl,wang2020directshape}. However, the availability of a suitable 3D prior is a rather strong assumption.

An alternative to address the ill-posedness of single-view reconstruction is to consider multiple views.
Recent methods for multi-view reconstruction assume camera parameters and use self-supervised learning  based on a neural renderer to reconstruct static and dynamic objects with novel 3D representations~\cite{mildenhall2020nerf,park2020nerfies}.
A downside of multi-view reconstruction methods is that they require many different images of the same object, which is often unavailable in existing datasets.

Contrary to existing approaches, we simultaneously solve deep multi-graph matching and infer sparse 3D geometry from inhomogeneous 2D image collections. Our approach obtains cycle-consistent multi-matchings and does not rely on a hand-crafted template or any other prior 3D model. 
\section*{Problem Formulation \& Preliminaries}
In this section we summarise how 
to achieve cycle-consistency for multiple graph matching by utilising the notion of universe points.
In order to explicitly construct such universe points, we consider the sparse reconstruction of 3D key points from multiple 2D images.

\subsection{Multi-Matching and Cycle Consistency}
Given is the set $\{ \cG_j\}_{j=1}^N$ of $N$ undirected graphs,
where each graph $\cG_j = (\cV_j, \cE_j)$ comprises of a total of $m_j$ nodes $\cV_j = \{v_1,  \dots, v_{m_j}\}$ and $n_j$ edges $\cE_j = \{e_1, \dots, e_{n_j}\}$ that connect pairs of nodes in $\cV_j$. We assume that each node represents an image key point, and that the node $v_i \in \bR^{2}$ is identified with the respective 2D image coordinates.
The pairwise graph matching problem is to find a node correspondence $X_{jk} \in \bP_{m_jm_k}$ between $\cG_j$ and $\cG_k$.
Here, $\bP_{m_jm_k}$ is the set of $(m_j {\times} m_k)$-dimensional partial permutation matrices.

Let $\cX = \{X_{jk} \in \bP_{m_jm_k} \}_{j,k=1}^N$ be the set of pairwise matchings between all graphs in $\{ \cG_j \}_{j=1}^N$.
$\cX$ is said to be cycle-consistent if for all $j, k, l \in \{1, \dots, N\}$, the following properties hold~\cite{huang2013consistent,Tron17,bernard2018nmfsync}:
\begin{enumerate}
    \item $X_{jj} = \mathbb{I}_{m_j}$, with the $m_j {\times} m_j$ identity matrix $\mathbb{I}_{m_j}$.
    \item $X_{jk} = X_{kj}^T$.
    \item $X_{jk}X_{kl} \leq X_{jl}$ (element-wise comparison).
\end{enumerate}
When solving multi-graph matchings with pairwise matching, cycle consistency is desirable since it is an intrinsic property of the (typically unknown) ground truth matching.
Rather then explicitly imposing the above three constraints, it is possible to achieve cycle consistency by representing the pairwise matching using a universe graph~\cite{huang2013consistent,Tron17,bernard2018nmfsync}:
\begin{lemma}
The set $\cX$ of pairwise matchings is cycle-consistent if there exists a collection $\{X_j \in \bP_{m_jd} : X_j \mathbf{1}_d = \mathbf{1}_{m_j}\}_{j=1}^N$ such that $\forall X_{jk} \in \cX$ it holds that $X_{jk} = X_j X_k^T$.
\end{lemma}
Here, the $X_j$ is the pairwise matching between the graph $\cG_j$ and a universe graph $\cU = (\cV, \cE)$ with $d$ universe points, where $\cV = \{u_1, \dots, u_d\}$ denote the universe points and $\cE = \{e_1, \dots, e_n\}$ the universe edges.
Intuitively, the universe graph can be interpreted as assigning each point in $\cG_j$  to one of the $d$ universe points in $\cU$.
Therefore, rather than modelling the cubic number of cycle consistency constraints on $\{\cG_j\}_{j=1}^N$ explicitly, we use an object-to-universe matching formulation based on the $\{X_j\}_{j=1}^N$.

\subsection{3D Reconstruction}\label{sec_3d_recon}
 Though the idea of the universe graph is a crucial ingredient for synchronisation approaches~\cite{pachauri2013permsync,huang2013consistent,bernard2018nmfsync}, the universe graph is  never explicitly instantiated in these methods. 
That is because it is merely used as an abstract entity that must exist in order to ensure cycle consistency in multi-matchings.
Considering that the graphs in this work come from image collections, we assume that the nodes $u_i \in \bR^3$ of the universe graph represent 3D points, which will allow us to address their explicit instantiation based on multiple-view geometry. 

We denote the homogeneous coordinate representation of the universe point $u_i \in \bR^3$ (represented in world coordinates) as $U_i = (u_i, 1) \in \bR^4$.
Its projection onto the $j$-th image plane, denoted by $\cV_{ij} = (v_{ij}, 1) \in \bR^3$, is given by
\begin{equation}
\cV_{ij} = 
\lambda_{ij}K_j 
\underbrace{\left(\begin{matrix}
1 & 0 & 0 & 0\\
0 & 1 & 0 & 0\\
0 & 0 & 1 & 0
\end{matrix} \right)}_{\Pi_0}
\underbrace{\left( \begin{matrix}
R_j & T_j \\
0 & 1
\end{matrix} \right)}_{g_j} U_i.
\label{eq:point_projection}
\end{equation}
Here, $g_j$ is the world-to-camera space rigid-body transformation comprising of the rotation $R_j \in \bR^{3 \times 3}$ and the translation $T_j \in \bR^3$, $\Pi_0$ is the canonical projection matrix, $K_j \in \bR^{3 \times 3}$ is the intrinsic camera matrix, and $\lambda_{ij} \in \bR$ is the scale parameter.
For  brevity, we define the general projection matrix $\Pi_j = K_j \Pi_0 g_j$.
Let $U \in \bR^{4\times d}$ be the stacked universe points in homogeneous coordinates, $\cV_j \in \bR^{3 \times d}$ be the respective projection onto the $j$-th image plane, and $\Lambda_j = \diag(\lambda_{1j},\ldots,\lambda_{dj}) \in \bR^{d \times d}$ be the diagonal scale matrix.
The matrix formulation of Eq. \eqref{eq:point_projection} is
\begin{equation}\label{eq:matrixMvg}
    \cV_j = \Pi_j U \Lambda_j.
\end{equation}
Once we have a collection of $N$ images of different objects from the same category (not necessarily the same object instance,~e.g.~two images of different bicycles), reconstructing the universe points $U$ can be phrased as solving Eq.~\eqref{eq:matrixMvg} in the least-squares sense, which reads
\begin{equation}
\begin{aligned}
    \argmin_{U} \sum_{j=1}^{N}
    || \Pi_j U\Lambda_j - \cV_j||_F^2.
    \label{eq:recon_3D}
    \end{aligned}
\end{equation}
Note that in practice the variables $U, \{\Lambda_j\}$ and $\{\Pi_j\}$ are generally unknown, so that without further constraints this is an under-constrained problem.
In the next section, we will elaborate on how we approach this.
\begin{figure*}[t!]
    \centering
    \includegraphics[width=\linewidth]{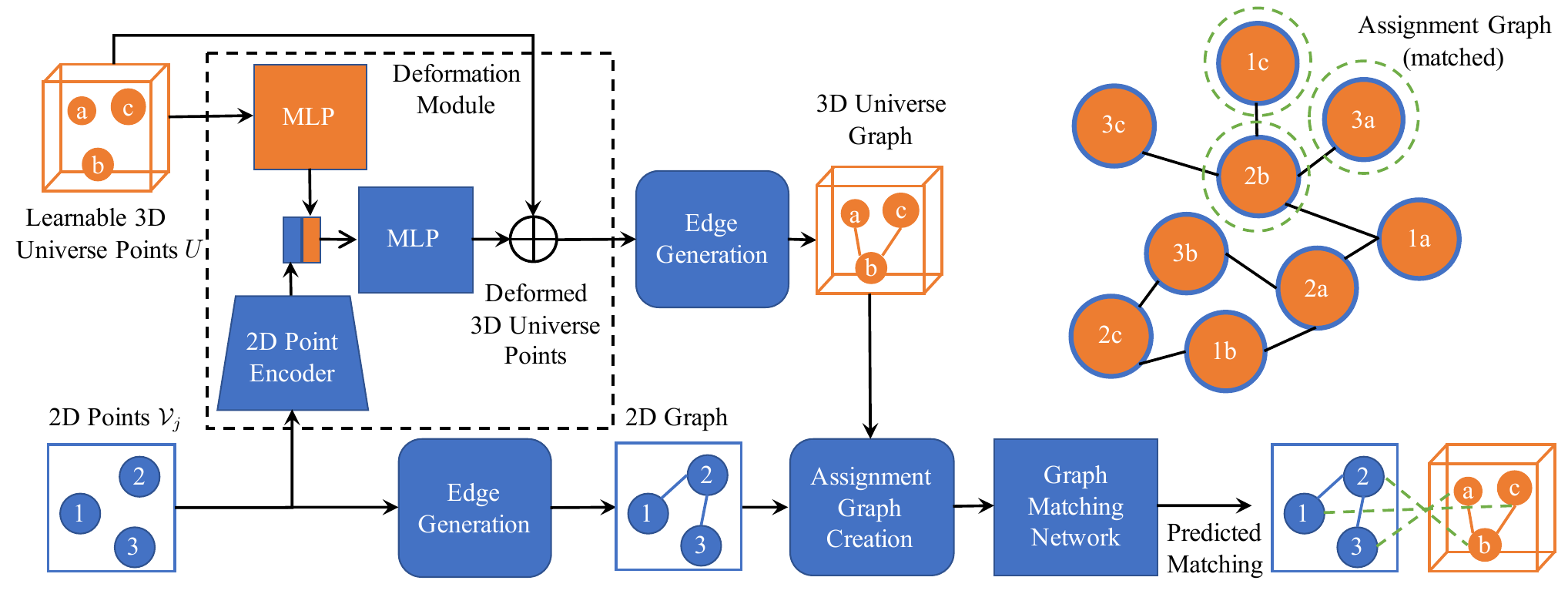}
    \caption{Overview of our algorithm. Given an image with 2D key points, we infer the corresponding image-specific 3D points in terms of a deformation of 3D universe points. The universe 3D points are learned during training for a given class of objects, while the deformations are predicted per image. We create edges and find a matching between the two graphs using a graph matching network. Since the matchings are between universe points and images, our matchings are intrinsically cycle consistent.}
    \label{fig:pipeline}
\end{figure*}

\section*{Proposed Method}
Our learning framework consists of four main components. The first two components have the purpose to obtain
 3D universe points, along with a deformation of these 3D points representing the underlying 3D structure of the 2D key points in the $j$-th image.
The purpose of the other two components is to predict the matching between the 2D points of $\cG_j$ and the 3D points of $\cU$. 
Thus, rather than learning  pairwise matchings between $\cG_j$ and $\cG_k$, we utilise an object-to-universe matching formulation. 
Therefore, the underlying 3D structure and cycle-consistent multi-matchings  are both attained by our method. 
The whole pipeline is illustrated in Fig.~\ref{fig:pipeline} and comprises the following four main components: 
\begin{enumerate}
    \item \textbf{Learnable 3D Universe Points}: the 2D key points $\{ \cV_j \}_{j=1}^N$ of all images in the collection are used to reconstruct the 3D universe points $U$ by incorporating a reconstruction loss that approximates Eq.~\eqref{eq:recon_3D}.
    \item \textbf{Deformation Module}: the retrieved universe points $U$ are static and therefore they cannot accurately model the geometric variability present in
     different instances of an object from the same category (e.g.~different bicycles).
    To address this, the universe points are non-linearly deformed by the
    deformation module that takes the 2D points and the (learned) 3D universe points as input. 

    \item \textbf{Assignment Graph Generation}: by connecting the 2D and universe points, respectively, the 2D graph and the 3D universe graph are constructed.
    The assignment graph is then constructed as the product of these two graphs.
    \item \textbf{Graph Matching Network}: a graph matching network performs graph convolutions on the assignment graph, and eventually performs a binary node classification on the assignment graph representing the matching between the 2D graph and the universe graph.

\end{enumerate}

\noindent\subsection{Learnable 3D Universe Points}
As discussed above,
the universe points can be retrieved by minimising~\eqref{eq:recon_3D}.
This problem, however, is generally under-determined, since $U, \{\Lambda_j\}$ and $\{\Pi_j\}$ in~\eqref{eq:recon_3D} are generally unknown in most practical settings.
Additionally, although all objects
share a similar 3D geometry, the nonlinear deformations between different instances are disregarded in \eqref{eq:recon_3D}. Thus, instead of an exact solution we settle for an approximation that we later refine in our pipeline. 
To this end, we assume a weak perspective projection model,~i.e.~all universe points are assumed to have the same distance from the camera.
With this condition, the diagonal of $\Lambda_j$ is constant and can be absorbed into 
$\Pi_j$.
This leads to the least-squares problem
\begin{equation}
    \argmin_U \sum_{j=1}^N || \Pi_j U - \cV_j||_F^2,
\end{equation}
which can be solved in an end-to-end manner during network training based on `backpropagable' pseudo-inverse implementations.
The variable $\Pi_j$ can be expressed as $\Pi_j = \cV_j U^+$, where $U^+$ is the right pseudo-inverse that satisfies $U U^+ = \mathbb{I}_4$.
Therefore, we solve the following problem
\begin{equation}
\label{eq:reconstructionLoss}
     U^* = \argmin_{U} \frac{1}{N} \sum_{j=1}^N || \cV_j U^+ U - \cV_j||^2_F.
\end{equation}

\subsection{Deformation Module}
The universe points retrieved in the previous step can only reflect 
the coarse geometric structure of the underlying 3D object, but cannot represent finer-scale variations between different instances within a particular object category. 
Thus, we introduce the deformation module to model an additional nonlinear deformation.

This module takes the universe points $U$ and the 2D points $\cV_j$ as input.
As shown in the bottom left of Fig.~\ref{fig:pipeline}, $\cV_j$ is passed to a 2D Point Encoder.
The encoder first performs a nonlinear feature transform of all input points based on multi-layer perceptron (MLP), and then performs a max pooling to get a global feature representing the input object.
As can be seen in the top left in Fig.~\ref{fig:pipeline}, an MLP is utilised  to perform a nonlinear feature transform for each of the 3D points in $U$.
Each 3D point feature is then concatenated with the same global feature from the 2D Point Encoder.
The concatenated per 3D point features are fed into an MLP 
to compute the deformation of each point.
The output is a set of per-point offsets $S \in \bR^{3 \times d}$ that
are added to $U$ to generate the deformed 3D universe points. 
The computation of the per-point offsets is summarised as 
\begin{equation}
\begin{aligned}
    S_j &= \text{MLP} \left( \text{MLP}(U) \circ \text{Encoder}(\cV_j)\right),
\end{aligned}
\end{equation}
where $\circ$ represents the concatenation operation.

 We enforce that the projection of the deformed universe points onto the image plane should be close to the observed 2D points, similar to the reconstruction loss in Eq. ~\eqref{eq:reconstructionLoss}.
Since the static 3D universe points should reflect the rough geometry of the underlying 3D object, the offset $S_j$ should be small.
Therefore, we introduce the deformed reconstruction loss and the offset regulariser as
    \begin{align}
    \cL_{\text{def}} &= \frac{1}{N} \sum_{j=1}^N ||\cV_j(U^*{+}S_j)^+(U^*{+}S_j) - \cV_j||^2_F, \text{ and}\\
    \cL_{\text{off}} &= ||S_j||^2_F.
    \end{align}

\subsection{Assignment Graph Generation}\label{sec:assignment}
To obtain graphs from the 2D points and the deformed 3D universe points, respectively, we utilise the Delaunay algorithm \cite{botsch2010polygon} to generate edges, see Fig.~\ref{fig:pipeline}. Moreover, we define the attribute of each edge as the concatenation of the  coordinates of the respective adjacent points. Note that other edge generation methods and attributes can be utilised as well. 

Once the 3D universe graph $\cU$ and the 2D graph $\cG_j$ are generated, we construct the assignment graph $\cG^A_j$ as the product graph of $\cU$ and $\cG_j$ following~\citet{leordeanu2005spectral}.
To be more specific, the nodes in $\cG^A_j$ are defined as the product of the two node sets $\cV_j$ (of $\cG_j$)  and $\cV$  (of $\cU$), respectively,~i.e.~$\cV^A_j = \{v_{jk}: v_{jk}= (v_j, u_k) \in \cV_j \times \cV\}$.
The edges in $\cG^A_j$ are built between nodes $v_{jk}$, $v_{mn} \in \cV^A_j$ if and only if there is an edge between $v_j$ and $v_m$ in $\cE_j$, as well as between $u_k$ and $u_n$ in $\cE$.
The attribute of each node and edge in $\cG^A_j$ is again the concatenation of the attribute of corresponding nodes and edges in $\cG_i$ and $\cU$, respectively.

\subsection{Graph Matching Network}
The graph matching problem is converted to a binary classification problem on the assignment graph $\cG^A$. 
For example, an assignment graph is shown on the top right of Fig.~\ref{fig:pipeline}.
Classifying nodes $\{1c, 2b, 3a\}$ as positive equals to matching point $1$ to $c$, $2$ to $b$ and $3$ to $a$, where numeric nodes correspond to the 2D graph, and alphabetic nodes correspond the 3D universe graph.

The assignment graph is then passed to the graph matching network~\cite{wang2020combinatorial}. A latent representation is achieved by alternatingly applying edge convolutions and node convolutions.
The edge convolution assembles the attributes of the connected nodes, while the node convolution aggregates the information from its adjacent edges and updates the attributes of each node. The overall architecture is based on the graph network from~\citet{battaglia2018relational}.

\begin{figure*}[t]
  \centering
  \footnotesize
  \newcommand{\mywidth}{0.12\textwidth} 
  \newcolumntype{C}{ >{\centering\arraybackslash} m{0.02\textwidth} }
  \newcolumntype{X}{ >{\centering\arraybackslash} m{\mywidth} }
  \setlength\tabcolsep{0.5pt} 
  \def\arraystretch{1.0} 
  \begin{tabular}[t]{XXXXXXXX}
  \multicolumn{4}{c}{Car (Willow Dataset)} & 
  \multicolumn{4}{c}{Duck (Willow Dataset)}\\
  \includegraphics[width=\mywidth]{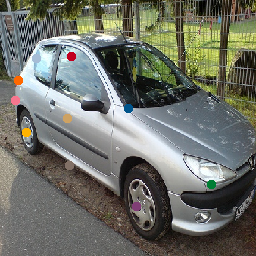} &
  \includegraphics[width=\mywidth]{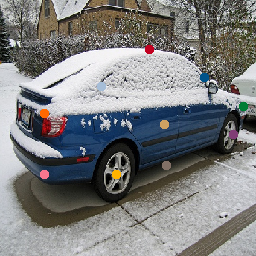} &
  \includegraphics[width=\mywidth]{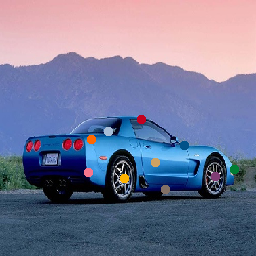} &
  \includegraphics[width=\mywidth]{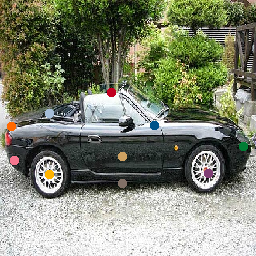} &
  \includegraphics[width=\mywidth]{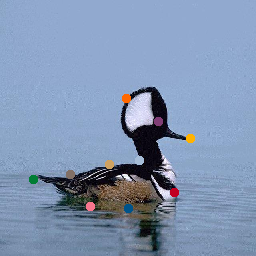} &
  \includegraphics[width=\mywidth]{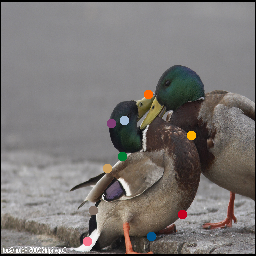} &
  \includegraphics[width=\mywidth]{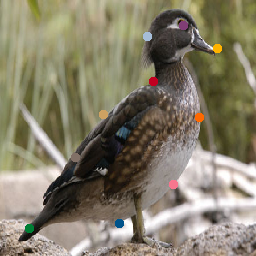} &
  \includegraphics[width=\mywidth]{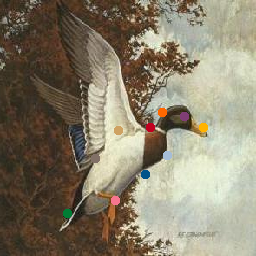} \\
  \multicolumn{4}{c}{Bicycle (Pascal VOC Dataset)} & 
  \multicolumn{4}{c}{Cow (Pascal VOC Dataset)} \\
  \includegraphics[width=\mywidth]{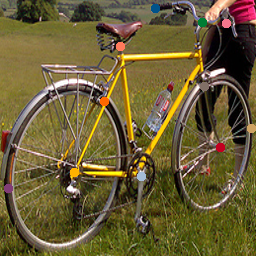} &
  \includegraphics[width=\mywidth]{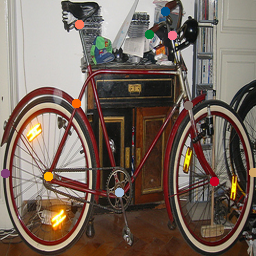} &
  \includegraphics[width=\mywidth]{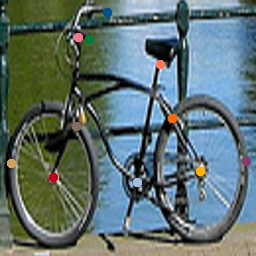} &
  \includegraphics[width=\mywidth]{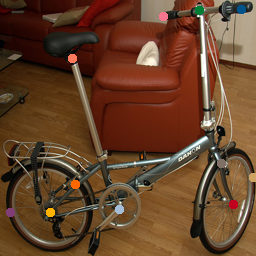} &
  \includegraphics[width=\mywidth]{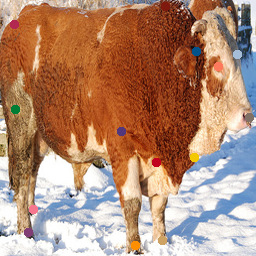} &
  \includegraphics[width=\mywidth]{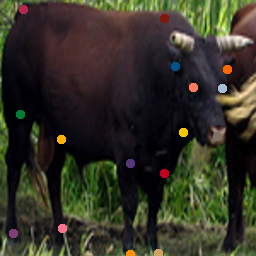} &
  \includegraphics[width=\mywidth]{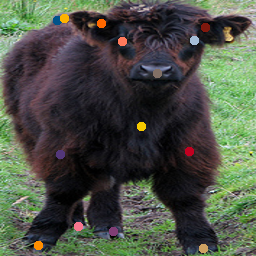} &
  \includegraphics[width=\mywidth]{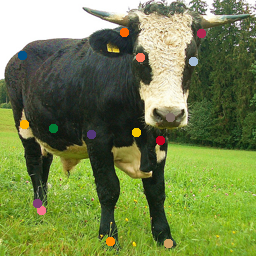}
  \end{tabular} 
  \caption{Qualitative results of our method on the Willow and Pascal VOC Dataset. We achieve accurate results for non-deformable objects of different types (car, bike) and reasonable results for instances of articulated objects (duck, cow).}
  \label{fig:qualitative_res}
\end{figure*}

\subsection{Loss Function}
Similarly as existing deep graph matching approaches, we train
our network in a supervised way based on the ground-truth matching matrix $X_j^{\text{gt}}$ between $\cG_j$ and $\cU$. To this end, we use the matching loss
\begin{equation}
    \cL_{\text{match}} = \frac{1}{N} \sum_{j=1}^N ||X_j^{\text{gt}} - X_j||^2_F.
\end{equation}
Furthermore, 
similarly as in previous work \cite{wang2018abpf, wang2020combinatorial}, we adopt a one-to-one matching prior in terms of a soft constraint.
To this end, we first convert the predicted permutation matrix $X_j$ to a binary node label matrix $Y_j \in \{0, 1\}^{m_jd \times 2}$ that we define as 
\begin{equation}
Y_j = \begin{pmatrix}1 {-} \text{vec}(X_{j}), \text{vec}(X_{j})\end{pmatrix}.
\end{equation}
Here, $\text{vec}(X_j)$ is the vectorisation of $X_j$.
We can compute the corresponding index vector $y_j \in \{0, 1\}^{m_jd}$ defined as
\begin{equation}
    (y_{j})_i = \argmax_{k \in \{1,2\}}(Y_{j})_{ik}.
\end{equation}
By leveraging the auxiliary matrix  $B \in \{0, 1\}^{(m_j+d) \times m_jd}$ and the ground-truth permutation matrix $X_j^{\text{gt}}$~\cite{wang2018abpf}, the one-to-one matching regularisation is 
\begin{equation}
    \cL_{\text{reg}} = || B (y - \text{vec}(X_j^{\text{gt}}))||^2.
\end{equation}
The total loss that we minimise during training is 
\begin{equation}
\begin{aligned}
    \cL = &~\omega_{\text{m}} \cL_{\text{match}} {+} \omega_{\text{d}} \cL_{\text{def}} {+}
     \omega_{\text{o}} \cL_{\text{off}} {+} \omega_{\text{reg}} \cL_{\text{reg}}.
    \end{aligned}
\end{equation}

\subsection{Training}
We train a single network that is able to handle multiple object categories at the same time. To this end, we learn separate 3D universe points for each category, and in addition  we introduce a separate learnable linear operator for each category that is applied to the global feature obtained by the 2D Point Encoder. The linear operator aims to transform the global feature to a category-specific representation, and also helps in resolving ambiguities between categories with objects that are somewhat similar (e.g.~cat and dog).

In practice, we apply a warm start to learn the universe points $\cU$, which are randomly initialised for each category. 
After retrieving $\cU$, we start training the neural network on the total loss with $\omega_{\text{m}} = 1$, $\omega_{\text{d}} = 0.5$, $\omega_{\text{o}} = 0.05$ and $\omega_{\text{reg}}=0.1$ (in all our experiments).
The batch size  is 16 and the number of iterations after warm start is 150k. 
The learning rate is $0.008$ and scheduled to decrease exponentially by 0.98 after each 3k iterations.
\section*{Experiments}
In the following, we evaluate our method in various settings. We compare our method to different state-of-the-art methods on two datasets, and we evaluate our deformation module based on a dataset of 3D objects.

\subsection{Ablation Study}
To confirm the importance of the individual components of our approach we conducted an ablation study. To this end we evaluate the accuracy on the Pascal VOC dataset in cases where we omit individual terms of the loss function, omit the warm start for learning the universe points $\cU$, and  omit deformation module, see Table~\ref{tab:ablation_reg}.
When we omit the one-to-one matching regulariser by setting $\omega_\text{reg}$ to 0, the matching accuracy is depressed substantially.
When we do not conduct a warm start for finding initial universe points, the matching accuracy deteriorates.
Similarly, the matching accuracy lowers without the use of our deformation module.
Further, the offset regularisation and the deformed reconstruction loss can refine the universe points for each object, which brings a better matching accuracy as shown in the last two experiments. Overall, the accuracy is highest when using all components together.
\begin{table}[h!]
	\fontsize{10}{11}
	\selectfont
    \centering
    \begin{tabular}{lc}
\toprule
\textbf{Ablative setting} & \textbf{Average accuracy}\\
\midrule
$\omega_{\text{reg}} = 0$ & 58.11\\
$\text{w/o warm start}$ & 58.49\\
$\text{w/o deformation module}$ & 60.33\\
$\omega_{\text{o}} = 0$ & 64.19\\
$\omega_{\text{d}} = 0$ & 64.81\\
Ours & 67.1\\
\bottomrule
    \end{tabular}

    \caption{Matching accuracy on the Pascal VOC dataset with the variants on regularisation terms or training strategies.
    }
    \label{tab:ablation_reg}
\end{table}

\subsection{Comparisons to the state of the art}
For the comparison experiments, we follow the testing protocol that was used in CSGM~\cite{wang2020combinatorial}. 
While all competing methods predict pairwise matchings $X_{ij}$, our approach predicts object-to-universe matchings $X_i$. Hence, we present the accuracies for pairwise matchings (written in parentheses) in addition to the accuracies for our object-to-universe matchings.
Note that $X_{ij}$ is obtained by $X_{ij} = X_i X_j^T$, which may add  individual errors in $X_i$ and $X_j$ up, thereby leading to smaller pairwise scores.
In the following, we summarise  the experimental setting for each dataset and discuss our results. Parts of the matching results are visualised in Fig.~\ref{fig:qualitative_res}.%
\setlength{\tabcolsep}{1.3pt}
\begin{table}[h!]
	\fontsize{10}{11}
	\selectfont
    \centering
    {
    \begin{tabular}{c|ccccc|ccc}
\toprule
 Method & car & duck & face & motor. & bottle & \textbf{Avg.} & \textbf{$\bigtriangleup$} & \textbf{3D}\\
\midrule
IPFP & 74.8 & 60.6 & 98.9 & 84.0 & 79.0 & 79.5 & \xmark & \xmark \\
RRWM & 86.3 & 75.5 & 100 & 94.9 & 94.3 & 90.2 & \xmark& \xmark \\
PSM & 88.0 & 76.8 & 100 & 96.4 & 97.0 & 91.6 & \xmark& \xmark \\
GNCCP & 86.4 & 77.4 & 100 & 95.6 & 95.7 & 91.0 & \xmark& \xmark \\
ABPF & 88.4 & 80.1 & 100 & 96.2 & 96.7 & 92.3 & \xmark& \xmark \\
HARG & 71.9 & 72.2 & 93.9 & 71.4 & 86.1 & 79.1 & \xmark& \xmark \\
GMN & 74.3 & 82.8 & 99.3 & 71.4 & 76.7 & 80.9 & \xmark& \xmark \\
PCA & 84.0 & 93.5 & 100 & 76.7 & 96.9 & 90.2 & \xmark& \xmark \\
CSGM & 91.2 & 86.2 & 100 & 99.4 & 97.9 & 94.9 & \xmark& \xmark \\
BBGM & 100.0 & 99.2 & 96.9 & 89.0 & 98.8 & 96.8 & \xmark & \xmark\\
Ours & 98.8 & 90.3 & 99.9 & 99.8 & 100 & 97.8 & \cmark& \cmark \\
Ours & (98.7) & (86.4) & (99.9) & (99.8) & (100) & (97.0) & \cmark& \cmark \\
\bottomrule
    \end{tabular}
    }
    \caption{Matching accuracy on Willow dataset, where `$\bigtriangleup$' indicates whether the method guarantees the cycle consistency, and `3D' indicates that 3D geometry is obtained. Comparing to the other algorithms, our method can achieve the best average accuracy and guarantee cycle consistency.
    }
    \label{tab:willowDataset}
\end{table}%
\begin{table}[h]
\fontsize{10}{11}
\selectfont
\centering
{
    \begin{tabular}{cc|ccc}
    
\toprule
Method & Filtering & \textbf{Avg. Acc.} & \textbf{$\bigtriangleup$} & \textbf{3D}\\
\midrule

GMN & y & 55.3 & \xmark & \xmark\\
PCA & y & 63.8 & \xmark & \xmark\\
CSGM & y & {68.5} & \xmark & \xmark\\
Ours & y & 67.1 & \cmark & \cmark\\
(Ours) & y & (58.9) & \cmark & \cmark\\ 
\midrule
BBGM-Max & n & 51.9 & \xmark & \xmark\\
BBGM & n & 61.4 & \xmark & \xmark\\
BBGM-Multi & n & 62.8 & locally & \xmark\\
Ours & n & 59.4 & \cmark & \cmark\\
(Ours) & n & (42.9) & \cmark & \cmark\\
\bottomrule
    \end{tabular}
}
    \caption{Results on Pascal VOC Keypoints dataset. Note that in terms of accuracy we achieve comparable results to the previous state of the art methods GMN \cite{zanfir2018gmn}, PCA~\cite{wang2019PCA}, CSGM~\cite{wang2020combinatorial} and BBGM~\cite{rolinek2020deep}, while we are the only one that additionally achieves cycle consistency (`$\bigtriangleup$') and reconstructs 3D geometry (`3D').
    }
    \label{tab:pascalvocDataset}
\end{table}%
\subsubsection{Willow Dataset}
\label{sec:willow}
We simultaneously train our model for all categories of the Willow dataset~\cite{cho2013HARG}. It consists of images from $5$ classes. It is compiled from Caltech-256 and Pascal VOC 2007 datasets, and consists of more than $40$ images per class with $10$ distinctively labelled features each.

We use the same training/testing split as in CSGM~\cite{wang2020combinatorial}. For training, $20$ images are randomly chosen from each class and the rest are used for testing. For non-learning based methods, the affinity matrix is constructed using the SIFT descriptors~\cite{lowe2004sift} as done by~\citet{wang2018abpf}, more details are described in supplementary material. We use the 2D key point coordinates as attributes of nodes in $\cG_i$, while the attributes of nodes in $\cU$ are the 3D coordinates of the (learned) universe points.

Table~\ref{tab:willowDataset} shows the accuracy of our method, on the Willow dataset, in comparison with IPFP~\cite{leordeanu2009ipfp}, RRWM~\cite{cho2010RRWM}, PSM~\cite{egozi2012PSM}, GNCCP~\cite{liu2013gnccp}, ABPF~\cite{wang2018abpf}, HARG~\cite{cho2013HARG}, GMN~\cite{zanfir2018gmn}, PCA~\cite{wang2019PCA},  CSGM~\cite{wang2020combinatorial} and BBGM~\cite{rolinek2020deep}. Our method achieves an average accuracy of $97.8\%$, while also being able to reconstruct the 3D structure of objects, see Fig.~\ref{fig:teaser}. In the car category, our method outperforms the others noticeably.
Although there is non-rigid motion  in the duck category caused by articulation, our method still achieve a reasonable accuracy.
Further, ours is the only one that guarantees  cycle-consistent matchings.

\subsubsection{Pascal VOC Keypoints Dataset}
The Pascal VOC Keypoints dataset~\cite{Bourdev2009pascalVocKey}
contains $20$ categories of objects with labelled key point annotations. The number of key points varies from 6 to 23 for each category. Following~\citet{wang2020combinatorial}, we use $7020$ images for training and $1682$ for testing.

\begin{figure}
    \centering
    \includegraphics[width=1.0\linewidth]{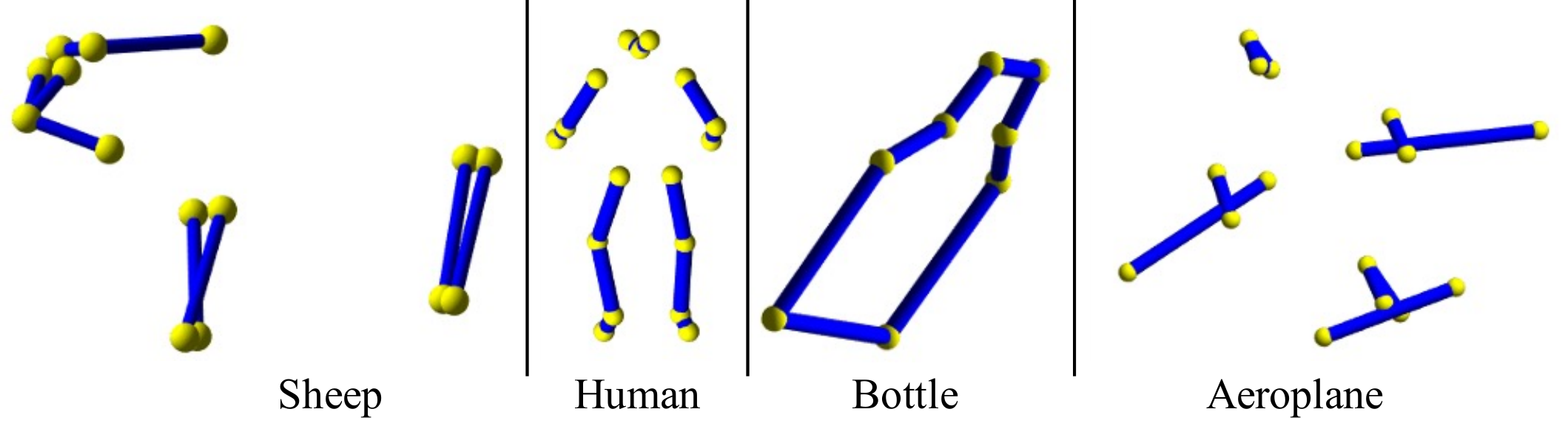}\\
    \caption{Illustration of 3D universe points. Examples of coarsev3D universe points from Pascal VOC dataset. Blue lines are handcrafted for better visualisation.}
    \label{fig:face_experiment}
\end{figure}

We randomly sample from the training data to train our model. As shown in Table~\ref{tab:pascalvocDataset}, in terms of matching accuracy our method is on par with the CSGM method.
Moreover, the ``Filtering'' column denotes that keypoints missing from one of the images are filtered out before matching. This procedure is not used for our method because the universe graph contains all possible key points in one category. Nevertheless, to provide a fair comparison in the ``Filtering'' setting, for our method we remove elements of the (non-binary) matching matrices corresponding to keypoints that are not presented, and binarize them afterwards. Furthermore, we also report accuracies for our method without any filtering.
Besides predicting accurate matchings, our method is the only one that achieves globally cycle-consistent matchings and infers 3D geometry as shown in Fig.~\ref{fig:face_experiment}. We emphasise that accuracy alone does not justifiably measure the performance of a method. Cycle consistency among the predicted matchings is also an important performance metric. More detailed results are provided in supp.~mat.

\subsection{3D Geometry and Deformation Evaluation} 
The goal of this experiment is to show that the learned 3D universe points are plausible, and the deformation module can compensate for instance-specific nonlinear deformations.
For this experiment, we use the 3D head dataset D3DFACs~\cite{cosker2012d3dfacs,li2017flame}.
We use a similar pre-processing pipeline as in i3DMM~\cite{yenamandra2020i3dmm} to obtain $8$ facial landmarks on each head in the template-registered dataset.
For training our model, we use 2D projections, with a pinhole camera model, of the randomly transformed 3D landmarks. During test time, we align the predicted 3D points with ground truth using Procrustes alignment to recover 3D scale and rigid transformation. The average L2 error between the ground truth 3D points and the obtained 3D universe points before and after deformations is $0.356$ and $0.148$, confirming the merits of the deformation module.
More qualitative results are provided in supp.~mat.
\section*{Conclusion}
In this work we tackle the novel problem setting of simultaneously solving graph matching and performing sparse 3D reconstruction from inhomogeneous 2D image collections. 
Our solution achieves several favourable properties simultaneously: our matchings are cycle-consistent, which is an important property since the (unknown) ground truth matchings are cycle-consistent. Our approach does not rely on the availability of an initial 3D geometry model, so that we can train it on virtually any object category, as opposed to object-specific 3D reconstruction approaches that are for example tailored towards faces only. Instead,
during training we learn a (sparse) deformable 3D geometric model directly from 2D image data. Moreover, our methods merely requires multiple images of \emph{different object instances} of the same category. This is in contrast to typical multi-view reconstruction approaches that require multiple images of the \emph{same object instance} from different views.
We believe that the joint consideration of deep graph matching and 3D geometry inference will open up interesting  research directions and that our approach may serve as inspiration for follow-up works on matching, 3D reconstruction, and shape model learning.

\bibliography{main}

\begin{thebibliography}{52}
\providecommand{\natexlab}[1]{#1}

\bibitem[{Battaglia et~al.(2018)Battaglia, Hamrick, Bapst, Sanchez-Gonzalez,
  Zambaldi, Malinowski, Tacchetti, Raposo, Santoro, Faulkner
  et~al.}]{battaglia2018relational}
Battaglia, P.~W.; Hamrick, J.~B.; Bapst, V.; Sanchez-Gonzalez, A.; Zambaldi,
  V.; Malinowski, M.; Tacchetti, A.; Raposo, D.; Santoro, A.; Faulkner, R.;
  et~al. 2018.
\newblock Relational Inductive Biases, Deep Learning, and Graph Networks.
\newblock \emph{arXiv preprint arXiv:1806.01261}.

\bibitem[{Bay et~al.(2008)Bay, Ess, Tuytelaars, and {Van Gool}}]{bay2008surf}
Bay, H.; Ess, A.; Tuytelaars, T.; and {Van Gool}, L. 2008.
\newblock Speeded-Up Robust Features (SURF).
\newblock \emph{Computer Vision and Image Understanding}.
\newblock Similarity Matching in Computer Vision and Multimedia.

\bibitem[{Bentley(1975)}]{bentley1975knn}
Bentley, J.~L. 1975.
\newblock Multidimensional Binary Search Trees Used for Associative Searching.
\newblock \emph{Communications of the ACM}.

\bibitem[{Bernard, Cremers, and Thunberg(2021)}]{bernard2021neurips}
Bernard, F.; Cremers, D.; and Thunberg, J. 2021.
\newblock Sparse Quadratic Optimisation over the Stiefel Manifold with
  Application to Permutation Synchronisation.
\newblock In \emph{NeurIPS}.

\bibitem[{{Bernard}, {Theobalt}, and {Moeller}(2018)}]{bernard2018dsstar}
{Bernard}, F.; {Theobalt}, C.; and {Moeller}, M. 2018.
\newblock DS*: Tighter Lifting-Free Convex Relaxations for Quadratic Matching
  Problems.
\newblock In \emph{Proceedings of the IEEE Conference on Computer Vision and
  Pattern Recognition}.

\bibitem[{Bernard et~al.(2018)Bernard, Thunberg, Goncalves, and
  Theobalt}]{bernard2018nmfsync}
Bernard, F.; Thunberg, J.; Goncalves, J.; and Theobalt, C. 2018.
\newblock Synchronisation of Partial Multi-Matchings via Non-negative
  Factorisations.
\newblock \emph{Pattern Recognition}.

\bibitem[{Bernard et~al.(2019)Bernard, Thunberg, Swoboda, and
  Theobalt}]{bernard19}
Bernard, F.; Thunberg, J.; Swoboda, P.; and Theobalt, C. 2019.
\newblock {HiPPI}: Higher-Order Projected Power Iterations for Scalable
  Multi-matching.
\newblock In \emph{Proceedings of the IEEE International Conference on Computer
  Vision}.

\bibitem[{Birdal and Simsekli(2019)}]{birdal2019probabilistic}
Birdal, T.; and Simsekli, U. 2019.
\newblock Probabilistic Permutation Synchronization using the Riemannian
  Structure of the Birkhoff Polytope.
\newblock In \emph{Proceedings of the IEEE Conference on Computer Vision and
  Pattern Recognition}.

\bibitem[{Bogo et~al.(2016)Bogo, Kanazawa, Lassner, Gehler, Romero, and
  Black}]{bogo2016keepitsmpl}
Bogo, F.; Kanazawa, A.; Lassner, C.; Gehler, P.; Romero, J.; and Black, M.~J.
  2016.
\newblock Keep it {SMPL}: Automatic Estimation of {3D} Human Pose and Shape
  from a Single Image.
\newblock In \emph{European Conference on Computer Vision}.

\bibitem[{Botsch et~al.(2010)Botsch, Kobbelt, Pauly, Alliez, and
  L{\'e}vy}]{botsch2010polygon}
Botsch, M.; Kobbelt, L.; Pauly, M.; Alliez, P.; and L{\'e}vy, B. 2010.
\newblock \emph{Polygon Mesh Processing}.
\newblock CRC press.

\bibitem[{Bourdev and Malik(2009)}]{Bourdev2009pascalVocKey}
Bourdev, L.~D.; and Malik, J. 2009.
\newblock Poselets: Body Part Detectors Trained using 3D Human Pose
  Annotations.
\newblock \emph{Proceedings of the IEEE International Conference on Computer
  Vision}.

\bibitem[{Burkard, Dell'Amico, and Martello(2012)}]{burkard2012}
Burkard, R.; Dell'Amico, M.; and Martello, S. 2012.
\newblock \emph{Assignment Problems: Revised Reprint}.
\newblock Society for Industrial and Applied Mathematics.

\bibitem[{Cho, Alahari, and Ponce(2013)}]{cho2013HARG}
Cho, M.; Alahari, K.; and Ponce, J. 2013.
\newblock Learning Graphs to Match.
\newblock In \emph{Proceedings of the IEEE International Conference on Computer
  Vision}.

\bibitem[{Cho, Lee, and Lee(2010)}]{cho2010RRWM}
Cho, M.; Lee, J.; and Lee, K.~M. 2010.
\newblock Reweighted Random Walks for Graph Matching.
\newblock In \emph{European Conference on Computer Vision}.

\bibitem[{Cosker, Krumhuber, and Hilton(2011)}]{cosker2012d3dfacs}
Cosker, D.; Krumhuber, E.; and Hilton, A. 2011.
\newblock A FACS Valid 3D Dynamic Action unit Database with Applications to 3D
  Dynamic Morphable Facial Modeling.
\newblock In \emph{Proceedings of the IEEE International Conference on Computer
  Vision}.

\bibitem[{Cour, Srinivasan, and Shi(2006)}]{Cour:2006un}
Cour, T.; Srinivasan, P.; and Shi, J. 2006.
\newblock {Balanced Graph Matching}.
\newblock \emph{Advances in Neural Information Processing Systems}.

\bibitem[{Dym, Maron, and Lipman(2017)}]{dym2017dspp}
Dym, N.; Maron, H.; and Lipman, Y. 2017.
\newblock DS++: A Flexible, Scalable and Provably Tight Relaxation for Matching
  Problems.
\newblock \emph{ACM Transactions on Graphics}.

\bibitem[{Egozi, Keller, and Guterman(2012)}]{egozi2012PSM}
Egozi, A.; Keller, Y.; and Guterman, H. 2012.
\newblock A Probabilistic Approach to Spectral Graph Matching.
\newblock \emph{IEEE Transactions on Pattern Analysis and Machine
  Intelligence}.

\bibitem[{Fey et~al.(2020)Fey, Lenssen, Morris, Masci, and
  Kriege}]{fey2020deep}
Fey, M.; Lenssen, J.~E.; Morris, C.; Masci, J.; and Kriege, N.~M. 2020.
\newblock Deep Graph Matching Consensus.
\newblock In \emph{International Conference on Learning Representations}.

\bibitem[{Huang and Guibas(2013)}]{huang2013consistent}
Huang, Q.-X.; and Guibas, L. 2013.
\newblock Consistent Shape Maps via Semidefinite Programming.
\newblock In \emph{Computer Graphics Forum}.

\bibitem[{Krizhevsky, Sutskever, and Hinton(2012)}]{krizhevsky2012alexnet}
Krizhevsky, A.; Sutskever, I.; and Hinton, G.~E. 2012.
\newblock ImageNet Classification with Deep Convolutional Neural Networks.
\newblock In \emph{Advances in Neural Information Processing Systems}.

\bibitem[{Lawler(1963)}]{lawler1963qap}
Lawler, E.~L. 1963.
\newblock The Quadratic Assignment Problem.
\newblock \emph{Management Science}.

\bibitem[{Leordeanu and Hebert(2005)}]{leordeanu2005spectral}
Leordeanu, M.; and Hebert, M. 2005.
\newblock A Spectral Technique for Correspondence Problems using Pairwise
  Constraints.
\newblock In \emph{IEEE International Conference on Computer Vision}.

\bibitem[{Leordeanu, Hebert, and Sukthankar(2009)}]{leordeanu2009ipfp}
Leordeanu, M.; Hebert, M.; and Sukthankar, R. 2009.
\newblock An Integer Projected Fixed Point Method for Graph Matching and Map
  Inference.
\newblock In \emph{Advances in Neural Information Processing Systems}.

\bibitem[{Li et~al.(2017)Li, Bolkart, Black, Li, and Romero}]{li2017flame}
Li, T.; Bolkart, T.; Black, M.~J.; Li, H.; and Romero, J. 2017.
\newblock Learning a Model of Facial Shape and Expression from {4D} Scans.
\newblock \emph{ACM Transactions on Graphics}.

\bibitem[{Li et~al.(2019)Li, Gu, Dullien, Vinyals, and Kohli}]{li2019graph}
Li, Y.; Gu, C.; Dullien, T.; Vinyals, O.; and Kohli, P. 2019.
\newblock Graph Matching Networks for Learning the Similarity of Graph
  Structured Objects.
\newblock In \emph{International Conference on Machine Learning}.

\bibitem[{Liu and Qiao(2013)}]{liu2013gnccp}
Liu, Z.-Y.; and Qiao, H. 2013.
\newblock GNCCP—Graduated Nonconvexity and Concavity Procedure.
\newblock \emph{IEEE Transactions on Pattern Analysis and Machine
  Intelligence}.

\bibitem[{Loiola et~al.(2007)Loiola, Abreu, Boaventura-Netto, Hahn, and
  Querido}]{loiola2007survey}
Loiola, E.; Abreu, N.; Boaventura-Netto, P.; Hahn, P.; and Querido, T. 2007.
\newblock A Survey of the Quadratic Assignment Problem.
\newblock \emph{European Journal of Operational Research}.

\bibitem[{Lowe(2004)}]{lowe2004sift}
Lowe, D. 2004.
\newblock Distinctive Image Features from Scale-Invariant Keypoints.
\newblock \emph{International Journal of Computer Vision}.

\bibitem[{Ma et~al.(2021)Ma, Jiang, Fan, Jiang, and Yan}]{ma2021image}
Ma, J.; Jiang, X.; Fan, A.; Jiang, J.; and Yan, J. 2021.
\newblock Image Matching from Handcrafted to Deep Features: A Survey.
\newblock \emph{International Journal of Computer Vision}.

\bibitem[{Maset, Arrigoni, and Fusiello(2017)}]{Maset:YO8y6VRb}
Maset, E.; Arrigoni, F.; and Fusiello, A. 2017.
\newblock {Practical and Efficient Multi-View Matching}.
\newblock In \emph{Proceedings of the IEEE International Conference on Computer
  Vision}.

\bibitem[{Mildenhall et~al.(2020)Mildenhall, Srinivasan, Tancik, Barron,
  Ramamoorthi, and Ng}]{mildenhall2020nerf}
Mildenhall, B.; Srinivasan, P.~P.; Tancik, M.; Barron, J.~T.; Ramamoorthi, R.;
  and Ng, R. 2020.
\newblock NeRF: Representing Scenes as Neural Radiance Fields for View
  Synthesis.
\newblock In \emph{European Conference on Computer Vision}.

\bibitem[{Pachauri, Kondor, and Singh(2013)}]{pachauri2013permsync}
Pachauri, D.; Kondor, R.; and Singh, V. 2013.
\newblock Solving the Multi-way Matching Problem by Permutation
  Synchronization.
\newblock In \emph{Advances in Neural Information Processing Systems}.

\bibitem[{Pardalos, Rendl, and Wolkowitz(1994)}]{pardalos1994quadratic}
Pardalos, P.; Rendl, F.; and Wolkowitz, H. 1994.
\newblock The Quadratic Assignment Problem: A Survey and Recent Developments.
  Quadratic Assignment and related problem.
\newblock \emph{DIMACS: Series in Discrete Mathematics and Theoretical Computer
  Science}.

\bibitem[{Park et~al.(2020)Park, Sinha, Barron, Bouaziz, Goldman, Seitz, and
  Martin-Brualla}]{park2020nerfies}
Park, K.; Sinha, U.; Barron, J.~T.; Bouaziz, S.; Goldman, D.~B.; Seitz, S.~M.;
  and Martin-Brualla, R. 2020.
\newblock Deformable Neural Radiance Fields.
\newblock \emph{arXiv preprint arXiv:2011.12948}.

\bibitem[{Rol{\'\i}nek et~al.(2020)Rol{\'\i}nek, Swoboda, Zietlow, Paulus,
  Musil, and Martius}]{rolinek2020deep}
Rol{\'\i}nek, M.; Swoboda, P.; Zietlow, D.; Paulus, A.; Musil, V.; and Martius,
  G. 2020.
\newblock Deep Graph Matching via Blackbox Differentiation of Combinatorial
  Solvers.
\newblock In \emph{European Conference on Computer Vision}.

\bibitem[{Swoboda et~al.(2017)Swoboda, Rother, Alhaija, Kainmüller, and
  Savchynskyy}]{swoboda2017b}
Swoboda, P.; Rother, C.; Alhaija, H.~A.; Kainmüller, D.; and Savchynskyy, B.
  2017.
\newblock Study of Lagrangean Decomposition and Dual Ascent Solvers for Graph
  Matching.
\newblock In \emph{Proceedings of the IEEE Conference on Computer Vision and
  Pattern Recognition}.

\bibitem[{Tewari et~al.(2017)Tewari, Zoll{\"o}fer, Kim, Garrido, Bernard,
  Perez, and Christian}]{tewari17MoFA}
Tewari, A.; Zoll{\"o}fer, M.; Kim, H.; Garrido, P.; Bernard, F.; Perez, P.; and
  Christian, T. 2017.
\newblock {MoFA: Model-based Deep Convolutional Face Autoencoder for
  Unsupervised Monocular Reconstruction}.
\newblock In \emph{Proceedings of the IEEE International Conference on Computer
  Vision}.

\bibitem[{Tron et~al.(2017)Tron, Zhou, Esteves, and Daniilidis}]{Tron17}
Tron, R.; Zhou, X.; Esteves, C.; and Daniilidis, K. 2017.
\newblock Fast Multi-Image Matching via Density-Based Clustering.
\newblock In \emph{Proceedings of the IEEE International Conference on Computer
  Vision}.

\bibitem[{Wang, Zhou, and Daniilidis(2018)}]{wang2018}
Wang, Q.; Zhou, X.; and Daniilidis, K. 2018.
\newblock Multi-Image Semantic Matching by Mining Consistent Features.
\newblock In \emph{Proceedings of the IEEE Conference on Computer Vision and
  Pattern Recognition}.

\bibitem[{Wang, Yan, and Yang(2019{\natexlab{a}})}]{wang2019PCA}
Wang, R.; Yan, J.; and Yang, X. 2019{\natexlab{a}}.
\newblock Learning Combinatorial Embedding Networks for Deep Graph Matching.
\newblock In \emph{Proceedings of the IEEE International Conference on Computer
  Vision}.

\bibitem[{Wang, Yan, and Yang(2019{\natexlab{b}})}]{wang2019neural}
Wang, R.; Yan, J.; and Yang, X. 2019{\natexlab{b}}.
\newblock Neural Graph Matching Network: Learning Lawler's Quadratic Assignment
  Problem with Extension to Hypergraph and Multiple-graph Matching.
\newblock \emph{arXiv preprint arXiv:1911.11308}.

\bibitem[{Wang, Yan, and Yang(2020)}]{wang2020Advances}
Wang, R.; Yan, J.; and Yang, X. 2020.
\newblock Graduated Assignment for Joint Multi-Graph Matching and Clustering
  with Application to Unsupervised Graph Matching Network Learning.
\newblock In Larochelle, H.; Ranzato, M.; Hadsell, R.; Balcan, M.~F.; and Lin,
  H., eds., \emph{Advances in Neural Information Processing Systems},
  volume~33, 19908--19919. Curran Associates, Inc.

\bibitem[{Wang et~al.(2020{\natexlab{a}})Wang, Yang, Stueckler, and
  Cremers}]{wang2020directshape}
Wang, R.; Yang, N.; Stueckler, J.; and Cremers, D. 2020{\natexlab{a}}.
\newblock DirectShape: Photometric Alignment of Shape Priors for Visual Vehicle
  Pose and Shape Estimation.
\newblock In \emph{Proceedings of the IEEE International Conference on Robotics
  and Automation}.

\bibitem[{Wang et~al.(2018)Wang, Ling, Lang, and Feng}]{wang2018abpf}
Wang, T.; Ling, H.; Lang, C.; and Feng, S. 2018.
\newblock Graph Matching with Adaptive and Branching Path Following.
\newblock \emph{IEEE Transactions on Pattern Analysis and Machine
  Intelligence}.

\bibitem[{Wang et~al.(2020{\natexlab{b}})Wang, Liu, Li, Jin, Hou, and
  Ling}]{wang2020combinatorial}
Wang, T.; Liu, H.; Li, Y.; Jin, Y.; Hou, X.; and Ling, H. 2020{\natexlab{b}}.
\newblock Learning Combinatorial Solver for Graph Matching.
\newblock In \emph{Proceedings of the IEEE Conference on Computer Vision and
  Pattern Recognition}.

\bibitem[{Yenamandra et~al.(2021)Yenamandra, Tewari, Bernard, Seidel, Elgharib,
  Cremers, and Theobalt}]{yenamandra2020i3dmm}
Yenamandra, T.; Tewari, A.; Bernard, F.; Seidel, H.; Elgharib, M.; Cremers, D.;
  and Theobalt, C. 2021.
\newblock i3DMM: Deep Implicit 3D Morphable Model of Human Heads.
\newblock In \emph{Proceedings of the IEEE Conference on Computer Vision and
  Pattern Recognition}.

\bibitem[{Zanfir and Sminchisescu(2018)}]{zanfir2018gmn}
Zanfir, A.; and Sminchisescu, C. 2018.
\newblock Deep Learning of Graph Matching.
\newblock In \emph{Proceedings of the IEEE Conference on Computer Vision and
  Pattern Recognition}.

\bibitem[{Zhang et~al.(2013)Zhang, Song, Shao, Zhao, and
  Shibasaki}]{zhang2013Oriented}
Zhang, Q.; Song, X.; Shao, X.; Zhao, H.; and Shibasaki, R. 2013.
\newblock Learning Graph Matching: Oriented to Category Modeling from Cluttered
  Scenes.
\newblock In \emph{Proceedings of the IEEE International Conference on Computer
  Vision}.

\bibitem[{Zhang and Lee(2019)}]{zhang2019deep}
Zhang, Z.; and Lee, W.~S. 2019.
\newblock Deep Graphical Feature Learning for the Feature Matching Problem.
\newblock In \emph{Proceedings of the IEEE International Conference on Computer
  Vision}.

\bibitem[{Zhou and De~la Torre(2016)}]{Zhou:2016ty}
Zhou, F.; and De~la Torre, F. 2016.
\newblock {Factorized Graph Matching}.
\newblock \emph{IEEE Transactions on Pattern Analysis and Machine
  Intelligence}.

\bibitem[{Zhou, Zhu, and Daniilidis(2015)}]{zhou2015multi}
Zhou, X.; Zhu, M.; and Daniilidis, K. 2015.
\newblock Multi-Image Matching via Fast Alternating Minimization.
\newblock In \emph{Proceedings of the IEEE International Conference on Computer
  Vision}.

\end{thebibliography}

\end{document}


\maketitle%

\section{Graph for non-learning based methods}
As the nodes are the key points in images, we need to construct the edges for each graph. Each edge $(k, l) \in \cE_j$ requires two features $w_{kl}$ and $\theta_{kl}$, where $w_{kl}$ is the pairwise distance between the connected nodes $v_k$ and $v_l$, and $\theta_{kl}$ is the absolute angle between the edge and the horizontal line with $0 \leq \theta_{kl} \leq \pi /2 $. The edge affinity between edges $(k, l)$ in $\cG_1$ and $(a,b)$ in $\cG_2$ is computed as $e_{(k,a), (l, b)} = \text{exp}(-(|w_{kl}-w_{ab}| + |\theta_{kl}-\theta_{ab}|)/2)$. The edge affinity can overcome the ambiguity of orientation because objects in real-world datasets typically have a natural up direction (e.g. people/animals stand on their feet, car/bikes on their tyres).

\begin{table*}[ht]
\footnotesize
\resizebox{\linewidth}{!}{
    \begin{tabular}{|c||c|c|c|c|c|c|c|c|c|c|c|c|c|c|c|c|c|c|c|c||c|}

\hline
 & aero & bike & bird & boat & bottle & bus & car & cat & chair & cow & table & dog & horse & mbike & person & plant & sheep & sofa & train & tv & Avg.\\
\hline

PCA & 40.92 & 15.48 & 44.91 & 45.30 & 14.55 & 41.83 & 55.97 & 42.97 & 35.99 & 44.30 & 41.59 & 49.10 & 43.68 & 33.33 & 35.04 & 24.67 & 53.93 & 45.87 & 44.00 & 29.39 & 39.19 \\
CSGM & 49.08 & 51.50 & 60.13 & 67.84 & 81.13 & 80.36 & 67.40 & 57.10 & 51.26 & 61.42 & 56.16 & 55.28 & 61.61 & 54.17 & 54.57 & 96.84 & 60.71 & 58.30 & 96.6 & 93.60 & 65.75 \\
Ours & 100 & 100 & 100 & 100 & 100 & 100 & 100 & 100 & 100 & 100 & 100 & 100 & 100 & 100 & 100 & 100 & 100 & 100 & 100 & 100 & 100\\
\hline
    \end{tabular}
}
    \vspace{1mm}
    \caption{Cycle consistency scores (in percent) of PCA~\cite{wang2019PCA}, CSGM~\cite{wang2020combinatorial} and ours on the Pascal VOC Keypoints dataset. Our method is the only one that guarantees cycle consistency for all categories.
    }
    \label{tab:pascalvoc_cycle_error}
\end{table*}

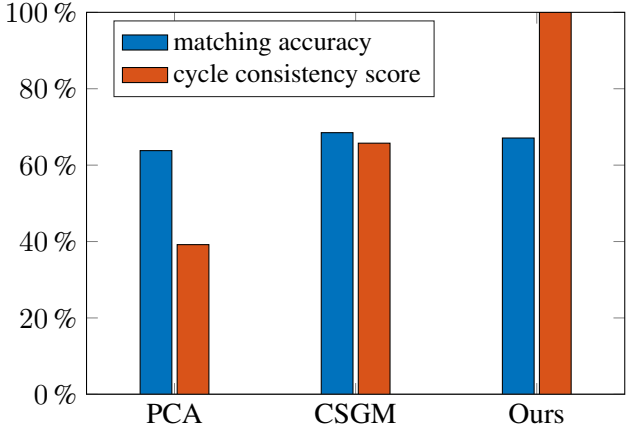
\begin{figure}
\definecolor{mycolor1}{rgb}{0.00000,0.44700,0.74100}%
\definecolor{mycolor2}{rgb}{0.85000,0.32500,0.09800}%
\newcommand{\barwidth}{12}
\begin{tikzpicture}

\begin{axis}[%
width=0.85\linewidth,
height=2in,
at={(0.05in,0.05in)},
scale only axis,
bar shift auto,
xmin=0.514285714285714,
xmax=3.48571428571429,
xtick={1,2,3},
xticklabels={{PCA},{CSGM},{Ours}},
ymin=0,
ymax=100,
axis background/.style={fill=white},
tick label style={font=\large},
yticklabel=\pgfmathparse{\tick}\pgfmathprintnumber{\pgfmathresult}\,\%,
legend style={at={(0.05,0.778)}, anchor=south west, legend cell align=left, align=left, draw=white!15!black}
]
\addplot[ybar, bar width=\barwidth, fill=mycolor1, draw=black, area legend] table[row sep=crcr] {%
1	63.8\\
2	68.5\\
3	67.1\\
};
\addplot[forget plot, color=white!15!black] table[row sep=crcr] {%
0.514285714285714	0\\
3.48571428571429	0\\
};
\addlegendentry{matching accuracy}

\addplot[ybar, bar width=\barwidth, fill=mycolor2, draw=black, area legend] table[row sep=crcr] {%
1	39.19\\
2	65.75\\
3	100\\
};
\addplot[forget plot, color=white!15!black] table[row sep=crcr] {%
0.514285714285714	0\\
3.48571428571429	0\\
};
\addlegendentry{cycle consistency score}

\end{axis}

\begin{axis}[%
width=5.833in,
height=1.375in,
at={(0in,0in)},
scale only axis,
xmin=0,
xmax=1,
ymin=0,
ymax=100,
axis line style={draw=none},
ticks=none,
axis x line*=bottom,
axis y line*=left,
legend style={legend cell align=left, align=left, draw=white!15!black}
]
\end{axis}
\end{tikzpicture}%
\caption{The average matching accuracy and cycle consistency score of PCA~\cite{wang2019PCA}, CSGM~\cite{wang2020combinatorial} and ours on Pascal VOC dataset.}
\label{fig:ma_cc}
\end{figure}

\section{Cycle Consistency}
We further provide quantitative evaluations of the cycle consistency on the Pascal VOC dataset, as shown in Table~\ref{tab:pascalvoc_cycle_error}.
We quantify in terms of the cycle consistency score, which is computed as follows:
\begin{enumerate}
    \item Given three graphs $\{\cG_j\}$, $\{\cG_k\}$ and $\{\cG_l\}$, we use the trained network to predict $X_{jk}$, $X_{jl}$ and $X_{kl}$.
    \item We compute the composed pairwise matching between $\{\cG_k\}$ and $\{\cG_l\}$ by $X^\prime_{kl} = X_{jk}^T X_{jl}$.\\
    \item We denote the number of points that $X^\prime_{kl}$ equals to $X_{kl}$ as $m_\text{cycle}$ and the number of points in $X_{kl}$ as $m_{kl}$. The cycle consistency score is then computed as 
    \begin{equation}
    \text{cycle consistency score} = 100 \times \frac{m_\text{cycle}}{m_{kl}} \%.
    \label{eq:cycle_error}
    \end{equation}
\end{enumerate}
Note that in this case, we only consider the common points that are observed in $\{\cG_j\}$, $\{\cG_k\}$ and $\{\cG_l\}$. 

In Fig.~\ref{fig:ma_cc}, we show the average matching accuracy and cycle consistency score of our method and compare it with PCA~\cite{wang2019PCA} and CSGM~\cite{wang2020combinatorial}. It is clear that our method can achieve comparable accuracy and the best cycle consistency at the same time.
\begin{figure*}
\includegraphics[width=\linewidth]{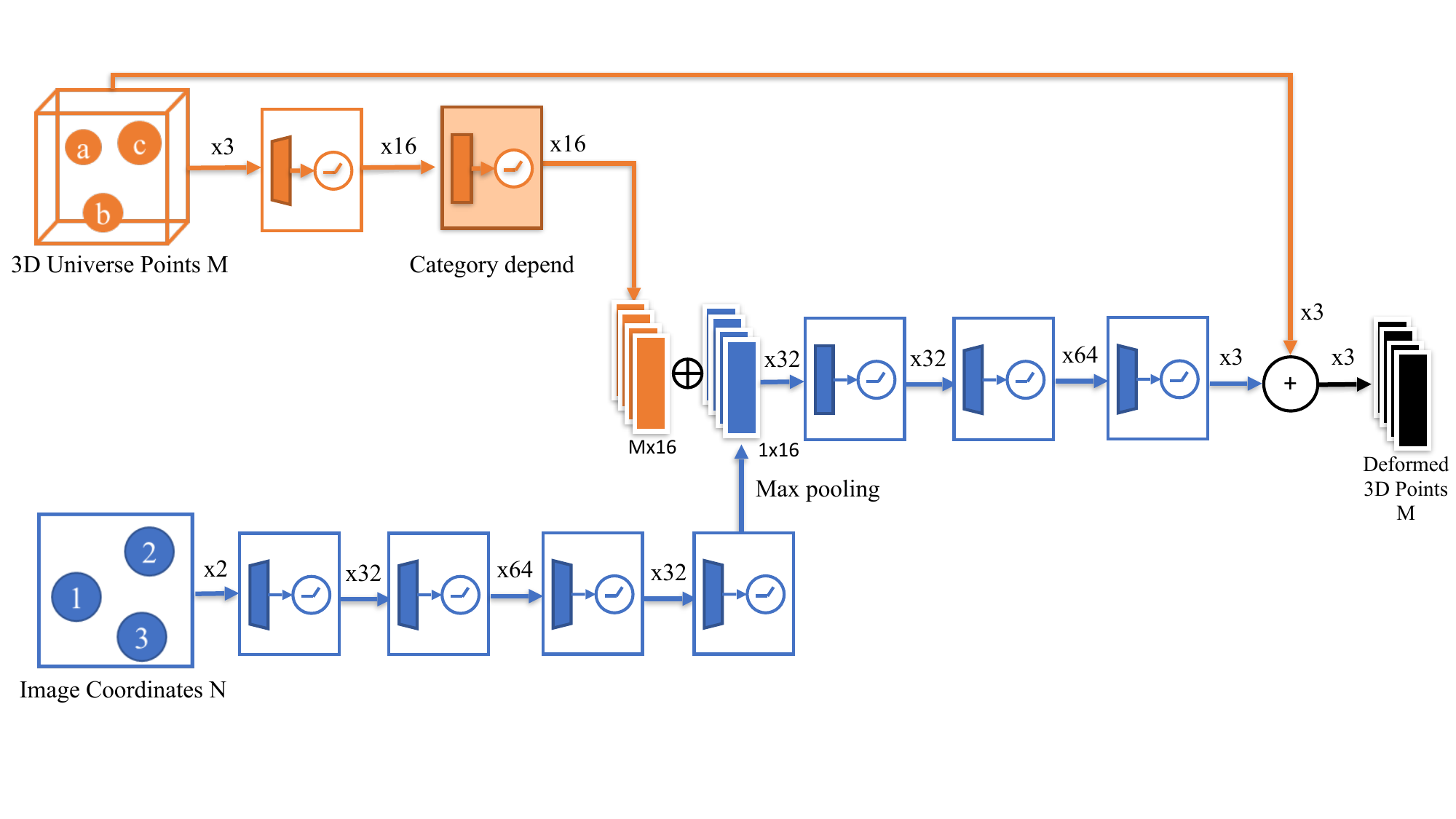}
\caption{The detailed architecture of the deformation module. Each box contains a linear layer and a ReLU unit. The linear layer in dashed box on 3D universe points is determined by the category of the input object. The goal of this layer is to alert the neural network about deformations in various categories.}
\label{fig:deform_arch}
\end{figure*}
\section{Network architecture}
We show the architecture of the deformation module in Fig.~\ref{fig:deform_arch}.  Each linear layer is followed by a Rectified Linear Unit (ReLU). Additionally, we introduce a linear layer depending on the category of the input object. Its purpose is to assist the neural network in distinguishing between different deformations among categories.
For detailed information on Graph Matching Network, readers are referred to \cite{wang2020combinatorial}

\section{More Deformation Results}
\begin{figure*}[ht]
  \centering
  \footnotesize
  \newcommand{\mywidth}{0.13\textwidth} 
  \newcolumntype{C}{ >{\centering\arraybackslash} m{0.05\textwidth} }
  \newcolumntype{X}{ >{\centering\arraybackslash} m{\mywidth} }
  \setlength\tabcolsep{0.5pt} 
  \def\arraystretch{1.0} 
  \begin{tabular}[ht]{CXXX||CXXX}
  $\cG$ & Right View & Front View & Left View &
  $\cG$ & Right View & Front View & Left View \\
  \rotatebox{90}{Ground Truth}&
  \includegraphics[width=\mywidth]{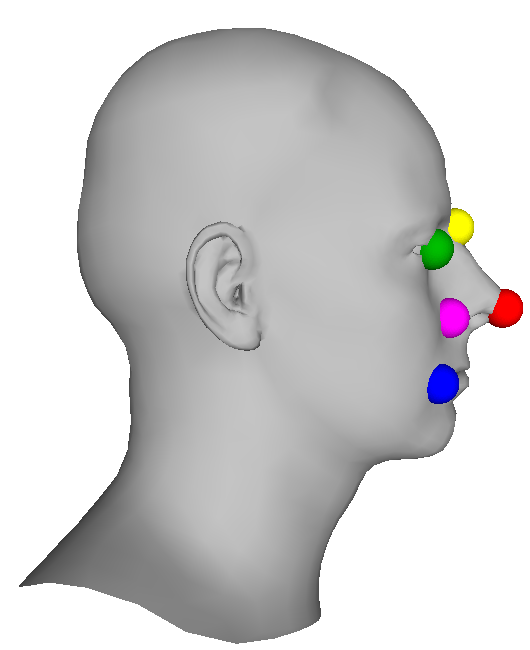} &
  \includegraphics[width=\mywidth]{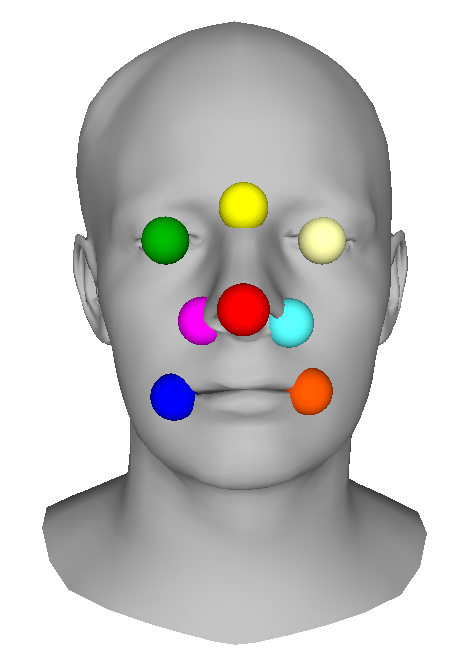} &
  \includegraphics[width=\mywidth]{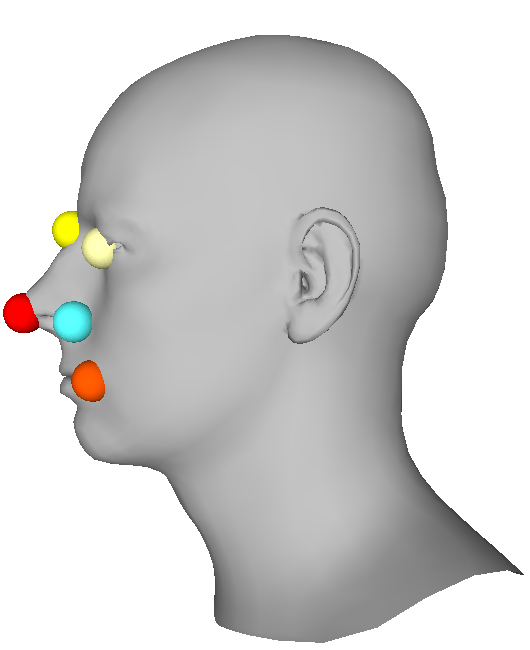} &
  \rotatebox{90}{Universe Points} &
  \includegraphics[width=\mywidth]{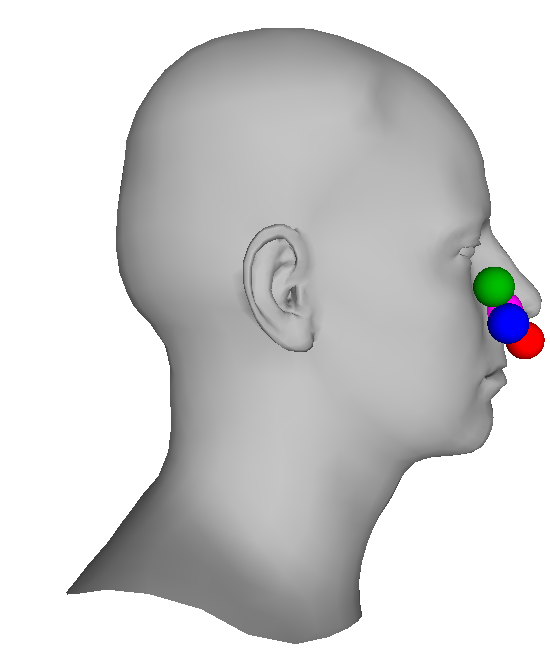} &
  \includegraphics[width=\mywidth]{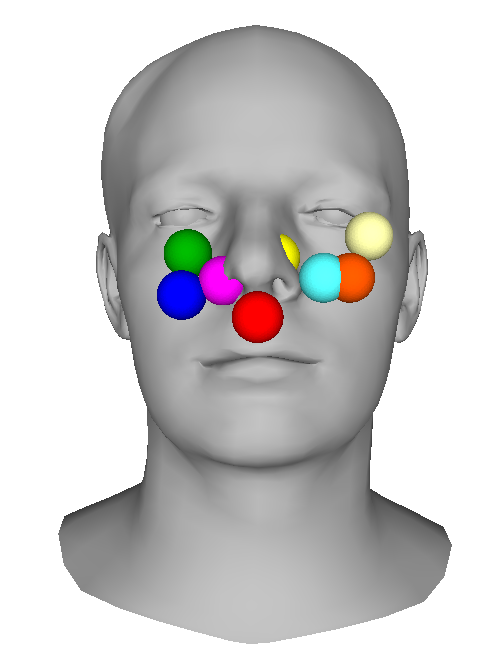} &
  \includegraphics[width=\mywidth]{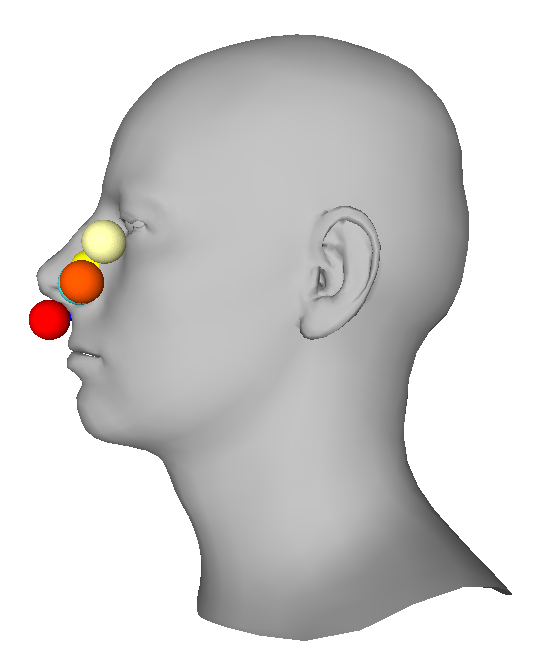} \\
  \hline 
  \hline
  \rotatebox{90}{Case 1} &
  \includegraphics[width=\mywidth]{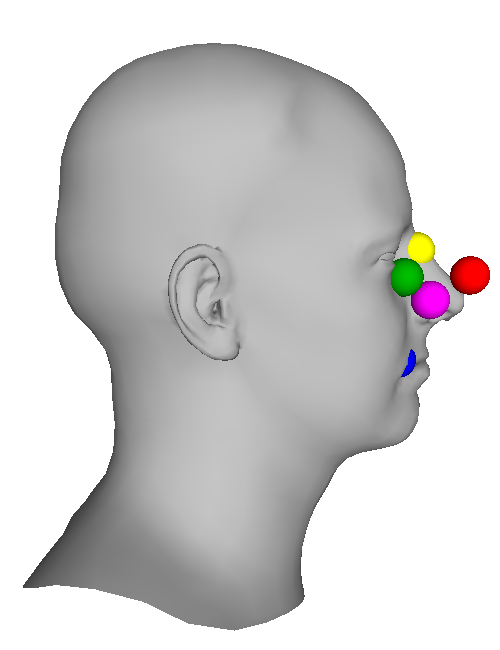} &
  \includegraphics[width=\mywidth]{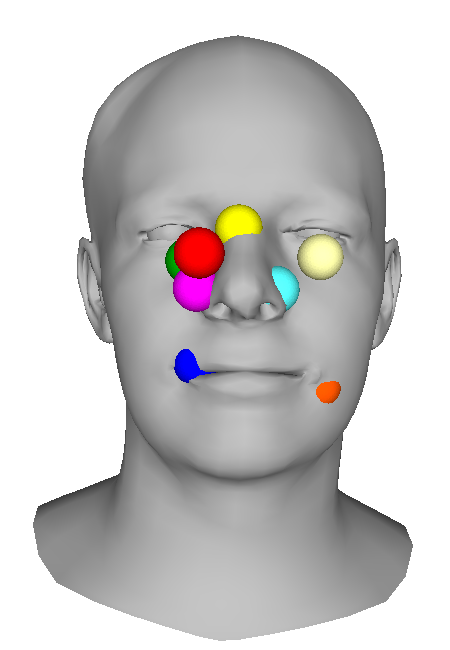} &
  \includegraphics[width=\mywidth]{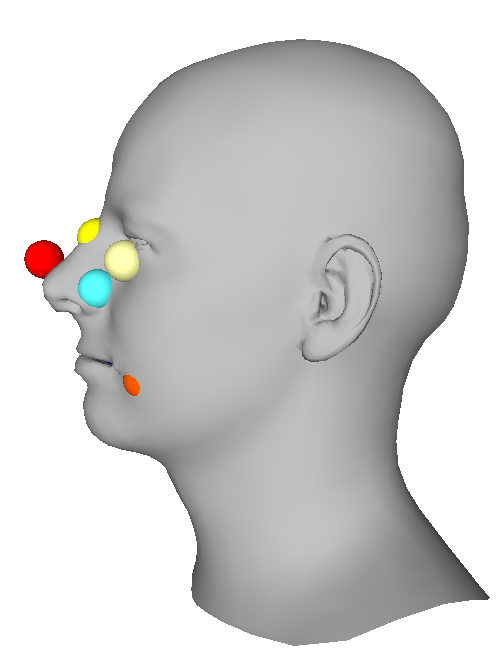} &
  \rotatebox{90}{Case 2} &
  \includegraphics[width=\mywidth]{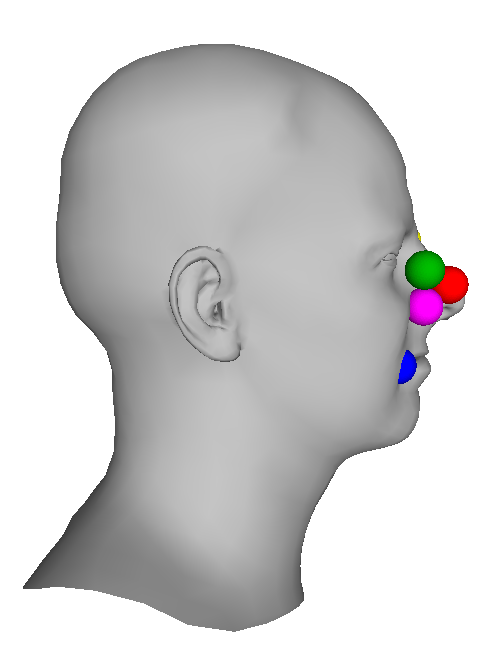} &
  \includegraphics[width=\mywidth]{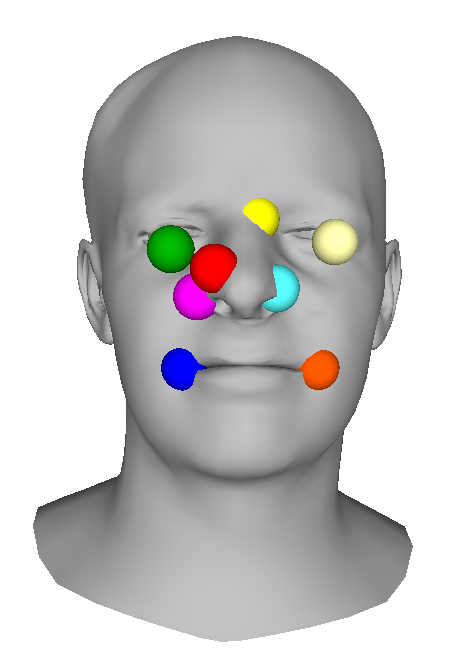} &
  \includegraphics[width=\mywidth]{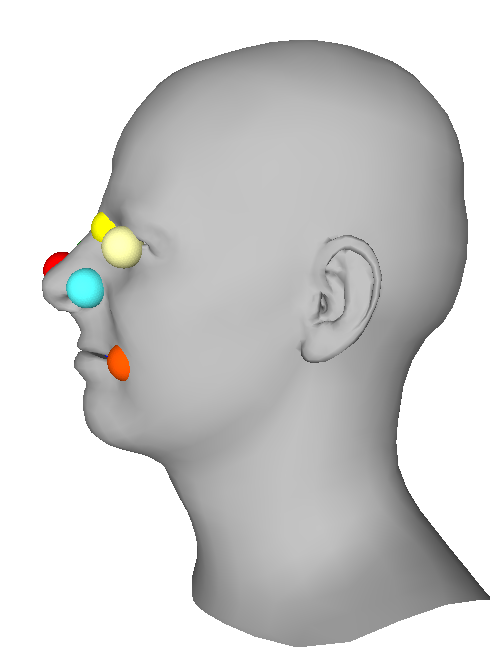} \\
  \rotatebox{90}{Case 3} &
  \includegraphics[width=\mywidth]{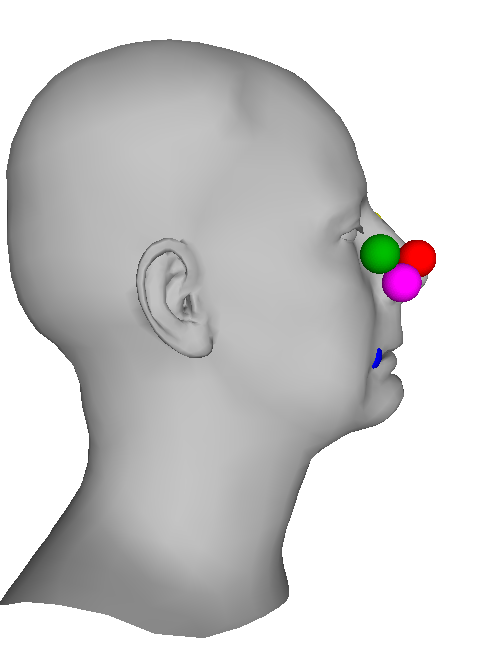} &
  \includegraphics[width=\mywidth]{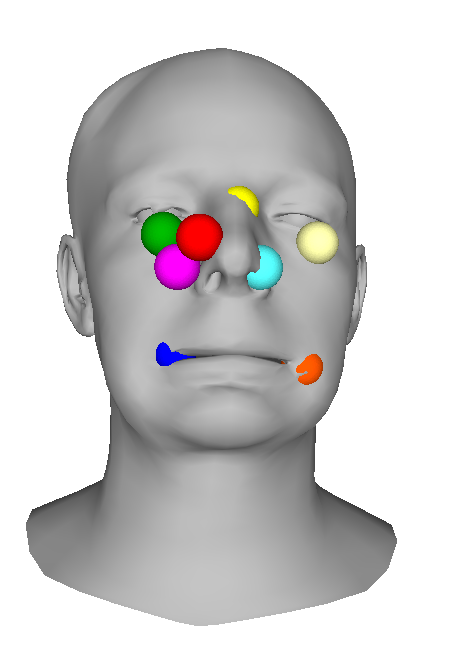} &
  \includegraphics[width=\mywidth]{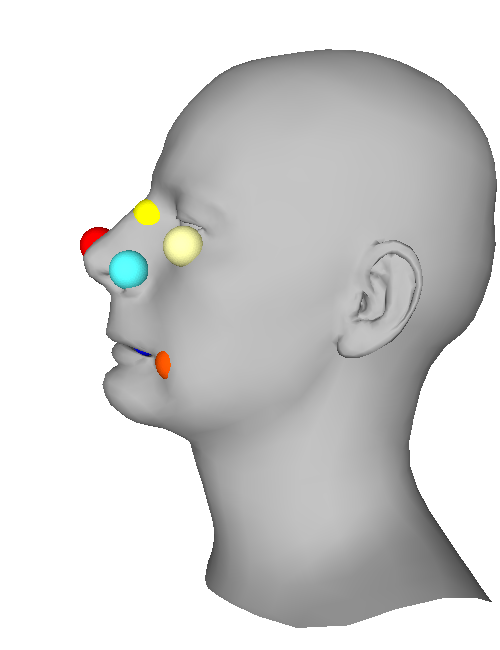} &
  \rotatebox{90}{Case 4} &
  \includegraphics[width=\mywidth]{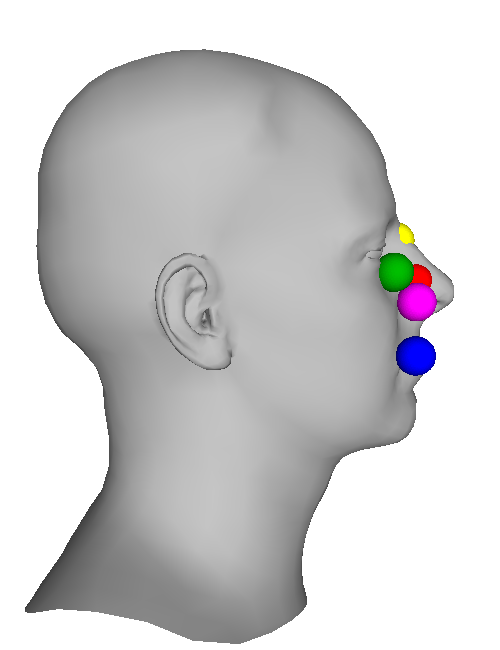} &
  \includegraphics[width=\mywidth]{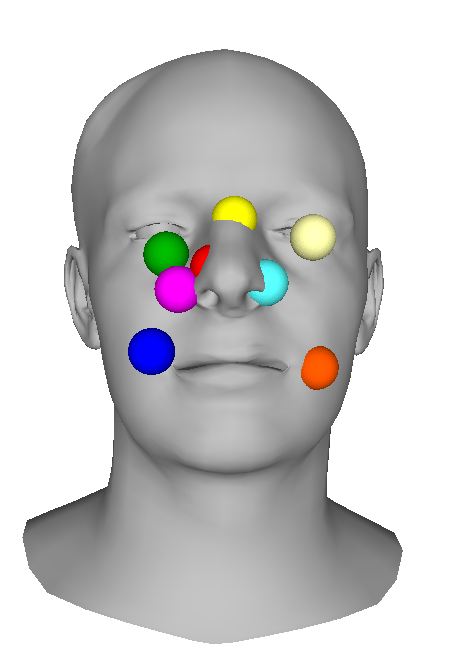} &
  \includegraphics[width=\mywidth]{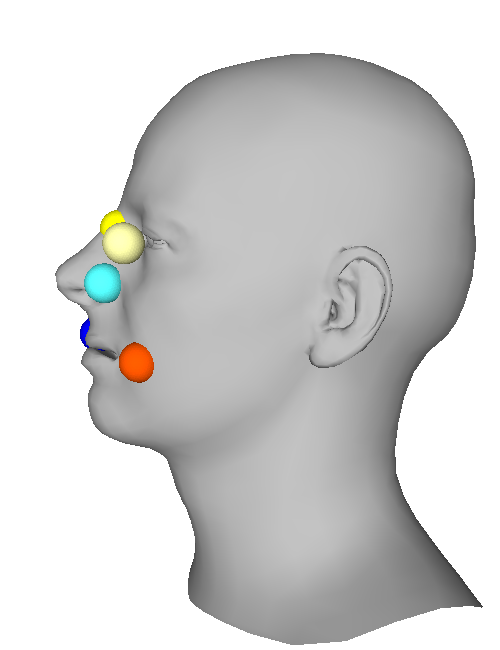} \\
  \rotatebox{90}{Case 5} &
  \includegraphics[width=\mywidth]{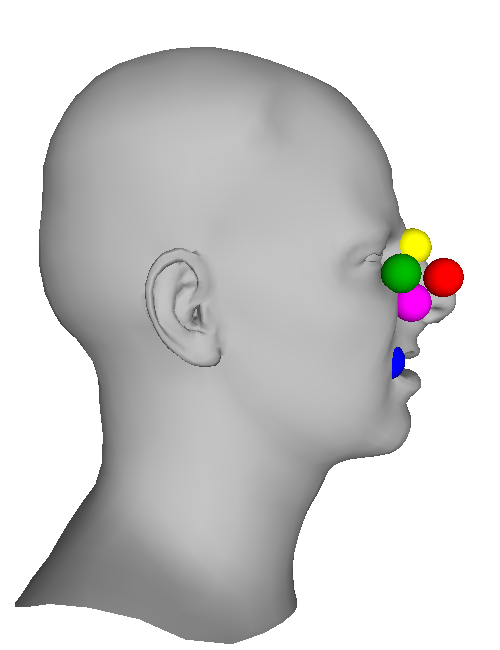} &
  \includegraphics[width=\mywidth]{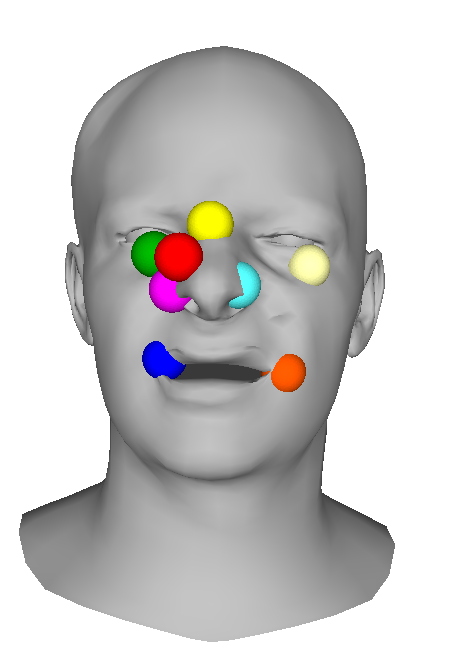} &
  \includegraphics[width=\mywidth]{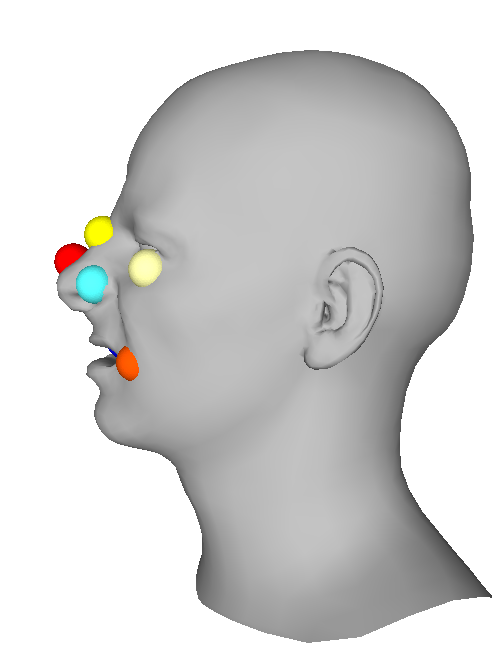} &
  \rotatebox{90}{Case 6} &
  \includegraphics[width=\mywidth]{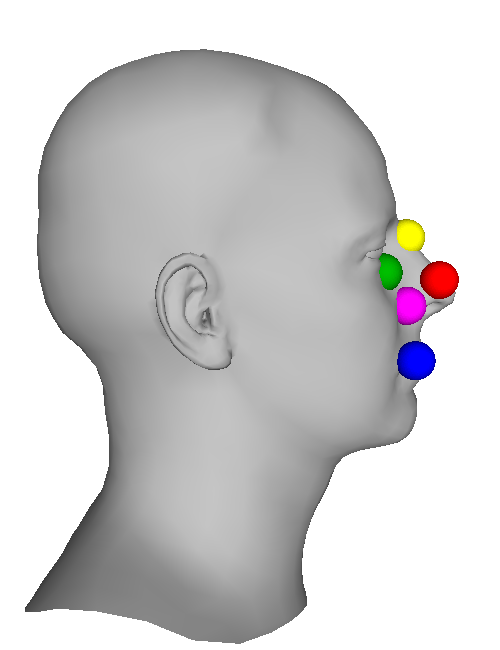} &
  \includegraphics[width=\mywidth]{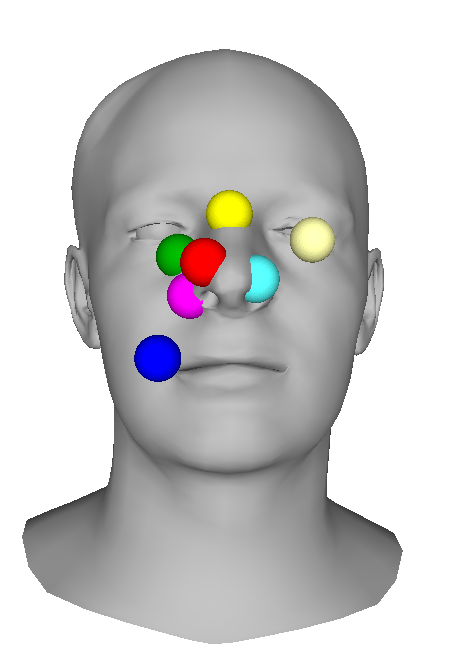} &
  \includegraphics[width=\mywidth]{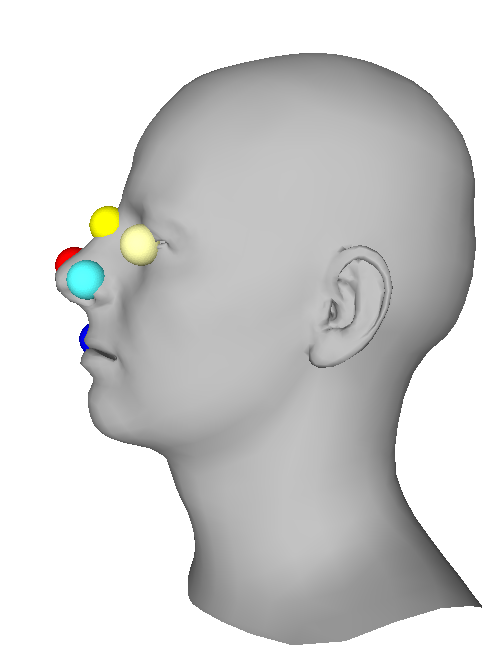} \\
  \rotatebox{90}{Case 7} &
  \includegraphics[width=\mywidth]{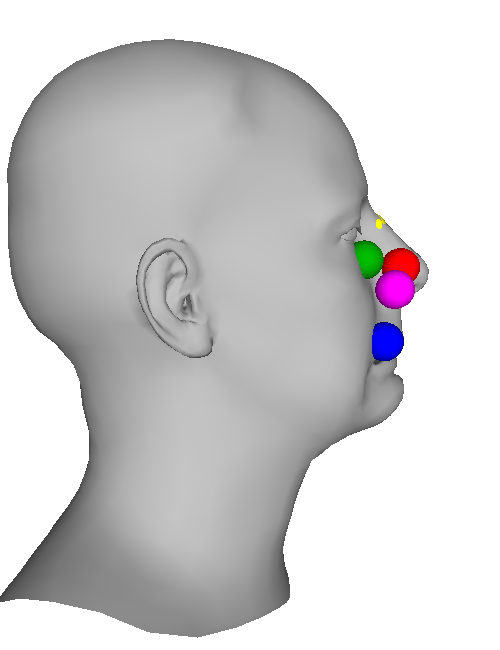} &
  \includegraphics[width=\mywidth]{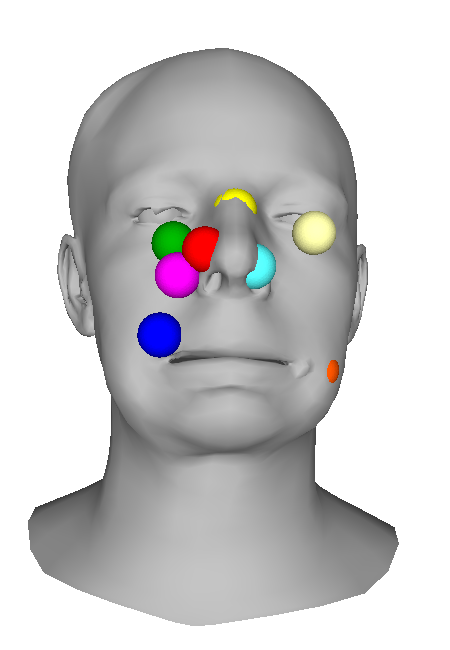} &
  \includegraphics[width=\mywidth]{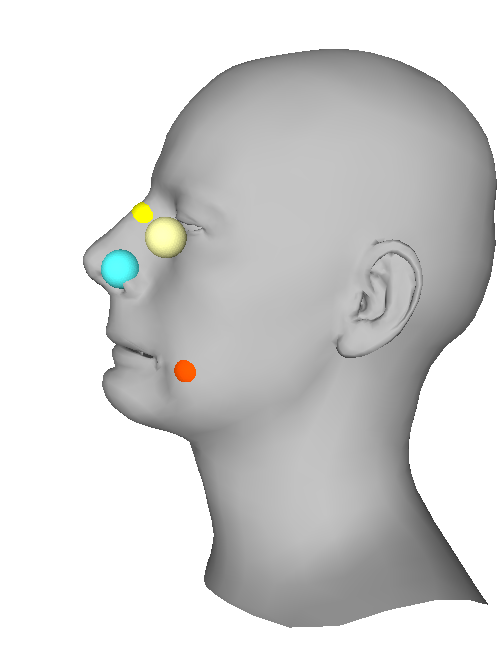} &
  \rotatebox{90}{Case 8} &
  \includegraphics[width=\mywidth]{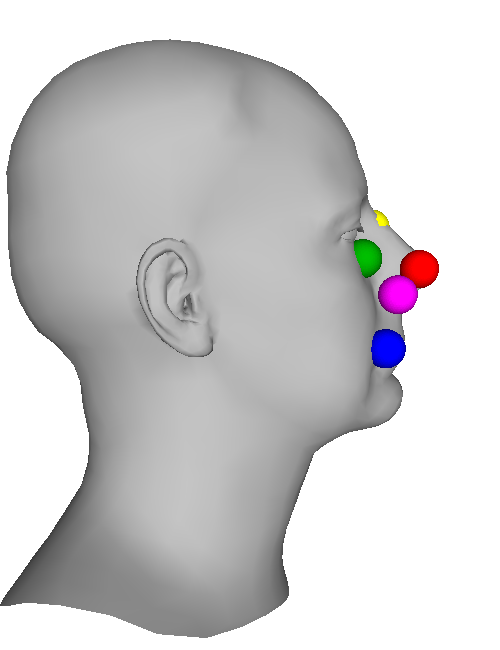} &
  \includegraphics[width=\mywidth]{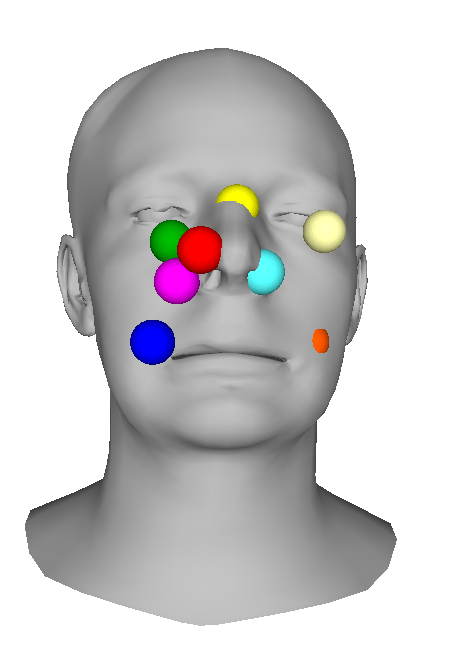} &
  \includegraphics[width=\mywidth]{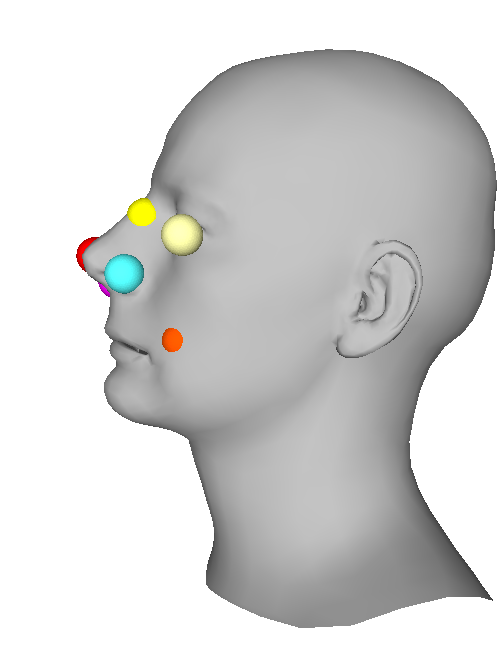} \\
  \end{tabular} 
  \caption{Qualitative results of deformation module. The top-left part shows the ground truth points on a reference shape, and the top-right part shows the universe points before the deformation module is applied.
  The remaining parts show individual cases, where it can be seen that the deformation module adequately deforms the universe points (top right), and that it is able to approximate the overall 3D geometry of the face well.}
  \label{fig:deformation_supp} \vspace{-0.4cm}
\end{figure*}

We provide more qualitative results for our deformation module, see Fig.~\ref{fig:deformation_supp}.
As shown in the figure, the deformation module is able to refine the 3D universe points. Although 3D reconstructions are not perfect, we can observe that they represent the overall 3D structure well, and are thus valuable for matching respective key points.

\bibliography{aaai22}